\documentclass[runningheads]{llncs}
\usepackage{graphicx}
\usepackage{amsmath,amssymb} 
\usepackage{url}
\usepackage{cite}
\usepackage{color}
\usepackage{epsfig}
\usepackage{graphicx}
\usepackage{wrapfig}
\usepackage{multirow}
\usepackage{tikz}
\usepackage{booktabs}
\usepackage{algorithmic}
\usepackage{arydshln}
\usepackage{colortbl}
\usepackage{enumitem}
\usepackage{siunitx}
\usepackage{bm}
\usepackage{array}
\usepackage{colortbl}
\usepackage{dblfloatfix}
\usepackage{marvosym}
\usepackage{lipsum}
\usepackage[title,toc,titletoc,page]{appendix}

\definecolor{mygray}{gray}{.9}
\definecolor{ggray}{RGB}{127,127,127}
\definecolor{reda}{RGB}{192,0,0}
\definecolor{redb}{RGB}{217,148,143}
\definecolor{myyellow}{RGB}{190,144,0}
\definecolor{mygreen}{RGB}{80,100,40}
\definecolor{myblue}{RGB}{30,90,100}

\newcommand{\etal}{\textit{et al}.}
\newcommand{\ie}{\textit{i}.\textit{e}.}
\newcommand{\eg}{\textit{e}.\textit{g}.}

\makeatletter
\newcommand{\thickhline}{%
    \noalign {\ifnum 0=`}\fi \hrule height 1pt
    \futurelet \reserved@a \@xhline
}

\usepackage[pagebackref=true,breaklinks=true,letterpaper=true,colorlinks,bookmarks=false]{hyperref}

\usepackage[width=122mm,left=12mm,paperwidth=146mm,height=193mm,top=12mm,paperheight=217mm]{geometry}

\begin{document}

\pagestyle{headings}
\mainmatter
\def\ECCVSubNumber{4046}  

\title{Active Visual Information Gathering for  Vision-Language Navigation} 

\titlerunning{Active Vision-Language Navigation}

\authorrunning{H. Wang, W. Wang, T. Shu, W. Liang, J. Shen}

\author{\small Hanqing Wang$^{1}$\and
\Letter Wenguan Wang$^{2}$\and
Tianmin Shu$^{3}$\and
Wei Liang$^{1}$\and
Jianbing Shen$^{4}$}
\institute{\small $^{1}$School of Computer Science, Beijing Institute of Technology~~$^{2}$ETH Zurich\\$^{3}$Massachusetts Institute of Technology~~$^{4}$Inception Institute of Artificial Intelligence\\
\small \url{https://github.com/HanqingWangAI/Active_VLN}}

\maketitle
\newcommand\blfootnote[1]{%
\begingroup
\renewcommand\thefootnote{}\footnote{#1}%
\addtocounter{footnote}{-1}%
\endgroup
}

\begin{abstract}
Vision-language navigation (VLN) is the task of entailing
an agent to carry out navigational instructions inside photo-realistic environments.
One of the key challenges in VLN is how to conduct a robust navigation by mitigating the uncertainty caused by ambiguous instructions and insufficient observation of the environment. Agents trained by current approaches typically suffer from this and would consequently struggle to avoid random and inefficient actions at every step. In contrast, when humans face such a challenge, they can still maintain robust navigation by actively exploring the surroundings to gather more information and thus make more confident navigation decisions. This work draws inspiration from human navigation behavior and endows an agent with an active information gathering ability for a more intelligent vision-language navigation policy. To achieve this, we propose an end-to-end framework for learning an exploration policy that decides \textbf{i)} when and where to explore, \textbf{ii)} what information is worth gathering during exploration, and \textbf{iii)} how to adjust the navigation decision after the exploration. The experimental results show promising exploration strategies emerged from training, which leads to significant boost in navigation performance. On the R2R challenge leaderboard, our agent gets promising results all three VLN settings, \ie, single run, pre-exploration, and beam search. 

\end{abstract}

\blfootnote{\Letter~Corresponding author: \textit{Wenguan Wang} (wenguanwang.ai@gmail.com).}

\section{Introduction}
Vision-language navigation (VLN)\!~\cite{anderson2018vision} aims to build an agent that can navigate a complex environment following human instructions. Existing methods have made amazing progress via \textbf{i)} efficient learning paradigms (\eg, using an ensemble of imitation learning and reinforcement learning\!~\cite{wang2018look,wang2019reinforced}, auxiliary task learning\!~\cite{wang2019reinforced,huang2019transferable,ma2019self,zhu2019vision}, or instruction augmentation based semi-supervised learning\!~\cite{fried2018speaker,tan2019learning}), \textbf{ii)} multi-modal information association\!~\cite{hu2019you}, and \textbf{iii)} self-correction\!~\cite{ke2019tactical,ma2019regretful}. However, these approaches have not addressed one of the core challenges in VLN -- the uncertainty caused by ambiguous instructions and partial observability.

\begin{figure}[t]
  \centering
      \includegraphics[width=0.99\linewidth]{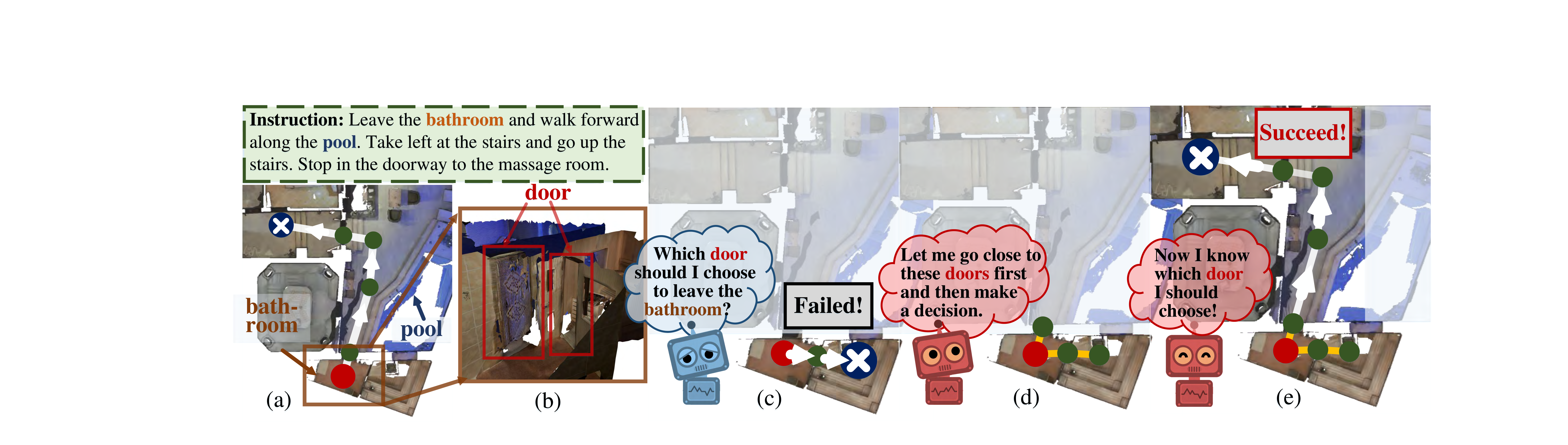}
\caption{{\!(a)\!~A top-down view of the environment with the groundtruth navigation path, based on the instructions. The start and end points are noted as red and blue circles, respectively. The navigation paths are labeled in white. (b) A side view of the bathroom in (a). (c) Previous agents face difficulties as there are two doors in the bathroom, hence causing the navigation fail. (d) Our agent is able to actively explore the environment for more efficient information collection. The exploration paths are labeled in yellow.  (e) After exploring the two doors, our agent executes the instructions successfully. 
}
}
\label{fig:overview}
\end{figure}

Consider the example in Fig.\!~\ref{fig:overview}, where an agent is required to navigate across rooms following human instructions: ``\textit{Leave the bathroom and walk forward along the pool.} $\cdots$''. The agent might be confused because the bathroom has two doors, and it consequently fails to navigate to the correct location (Fig.\!~\ref{fig:overview}(c)). In contrast, when faced with the same situation, our humans may perform better as we would  first explore the two doors, instead of directly making a \textit{risky} navigation decision. After collecting enough information, \ie,
confirming which one allows us to ``\textit{walk forward along the pool}'', we can take a more confident navigation action. This insight from human navigation behavior motivates us to develop an agent that has a similar active exploration and information gathering capability. When facing ambiguous instructions or low confidence on his navigation choices, our agent can actively explore his surroundings and gather information to better support navigation-decision making  (Fig.\!~\ref{fig:overview}(d-e)). However, previous agents are expected to conduct navigation at all times and only collect information from a limited scope. Compared with these, which perceive a scene \textit{passively}, our agent gains a larger \textit{visual field} and improved robustness against complex environments and ambiguous instructions by actively exploring the surrounding.

To achieve this, we develop an active exploration module, which learns to 1) decide when the exploration is necessary, 2) identify which part of the surroundings is worth exploring, and 3) gather useful knowledge from the environment to support more robust navigation. During training, we encourage the agent to collect relevant information to help itself make better decisions. We empirically show that our exploration module successfully learns a good information gathering policy and, as a result, the navigation performance is significantly improved.

With above designs, our agent gets promising results on R2R\!~\cite{anderson2018vision} benchmark leaderboard, over all three VLN settings, \ie, single run, pre-exploration, and beam search. In addition, the experiments show that our agent performs well in both seen and unseen environments.

\section{Related Work}
\noindent \textbf{Vision and Language.} Over the last few years, unprecedented advances in the design and optimization
of deep neural network architectures have led to tremendous progress  in computer vision and natural language processing. This progress, in turn, has enabled  a multitude of multi-modal applications spanning both
disciplines, including image captioning\!~\cite{xu2015show}, visual question answering\!~\cite{Antol_2015_ICCV}, visual grounding\!~\cite{yu2016modeling}, visual dialog\!~\cite{das2017visual,zheng2019reasoning}, and vision-language navigation\!~\cite{anderson2018vision}.  The formulation of these tasks requires a comprehensive understanding of both visual and linguistic content. A typical solution is to learn a joint multi-modal embedding space, \ie, CNN-based visual features and RNN-based linguistic representations are mapped to a common space by several non-linear operations. Recently, neural attention\!~\cite{xu2015show}, which is good at mining cross-modal knowledge, has shown to be a pivotal technique for multi-modal representation learning.

\noindent \textbf{Vision-Language Navigation  (VLN).} In contrast to previous vision-language tasks (\eg, image captioning, visual dialog) only involving \textit{static} visual content, VLN entails an agent to \textit{actively} interact with the environment to fulfill navigational instructions. Although VLN is relatively new in computer vision (dating back to\!~\cite{anderson2018vision}), many of its core units/technologies (such as instruction following\!~\cite{andreas2015alignment} and instruction-action mapping\!~\cite{mei2016listen}) were introduced much earlier. Specifically, these were originally studied in natural language processing and robotics communities, for the focus of either language-based navigation in a controlled environmental context\!~\cite{macmahon2006walk,tellex2011understanding,chen2011learning,andreas2015alignment,mei2016listen,misra2018mapping}, or vision-based navigation in visually-rich real-world scenes\!~\cite{mirowski2017learning,zhu2017target}. The VLN simulator described in\!~\cite{anderson2018vision} unites these two lines of research, providing photo-realistic environments and human-annotated instructions (as opposed
to many prior efforts using virtual scenes or formulaic instructions). Since its release, increased research has been conducted in this direction. Sequence-to-sequence\!~\cite{anderson2018vision} and reinforcement learning\!~\cite{wang2018look} based solutions were first adopted. Then, \cite{fried2018speaker,tan2019learning} strengthened the navigator by synthesizing new instructions. Later, combining imitation learning and reinforcement learning became a popular choice\!~\cite{wang2019reinforced}. Some recent studies explored auxiliary tasks as self-supervised signals\!~\cite{wang2019reinforced,huang2019transferable,ma2019self,zhu2019vision}, while some others addressed self-correction for intelligent path planning\!~\cite{ke2019tactical,ma2019regretful}. In addition, Thomason \etal\!~\cite{thomason2019shifting} identified unimodal biases in VLN, and Hu \etal\!~\cite{hu2019you} then achieved multi-modal grounding using a mixture-of-experts framework.

\section{Methodology}

\noindent \textbf{Problem Description.} Navigation in the Room-to-Room task\!~\cite{anderson2018vision} demands an agent to perform a sequence of navigation actions in real indoor environments and reach a target location by following natural language instructions.


\noindent \textbf{Problem Formulation and Basic Agent.} 
Formally, a language instruction is represented via textual embeddings as $\textbf{\textit{X}}$. At each navigation step $t$, the agent has a panoramic view\!~\cite{fried2018speaker}, which is discretized into 36 single views (\ie, RGB images). The agent makes a navigation decision in the panoramic action space, which consists of $K$ navigable views (reachable and visible), represented as $\!\textbf{\textit{V}}_{\!t}\!=\!\{\textbf{\textit{v}}_{t,1},\textbf{\textit{v}}_{t,2},\cdots\!,\textbf{\textit{v}}_{t,K}\}$. The agent needs to
make a decision on which navigable view to go to (\ie, choose an action $a^{\text{nv}\!}_t\!\in_{\!}\!\{1,\cdots\!,K\}$ with the embedding $\textbf{\textit{a}}^{\text{nv}\!}_t\!=_{\!}\!\textbf{\textit{v}}_{t,a^{\text{nv\!}}_t}$), according to the given instruction $\textbf{\textit{X}}$,  history panoramic views $\{\textbf{\textit{V}}_{\!1},\textbf{\textit{V}}_{\!2},\cdots\!,\textbf{\textit{V}}_{\!t-1}\}$ and previous actions $\{\textbf{\textit{a}}^{\text{nv}}_1,\textbf{\textit{a}}^{\text{nv}}_2,\cdots\!,\textbf{\textit{a}}^{\text{nv}}_{t-1}\}$.
Conventionally,  this dynamic navigation process is formulated in a recurrent form~\cite{anderson2018vision,tan2019learning}:
\begin{equation}
\begin{aligned}
\textbf{\textit{h}}^{\!\text{nv}}_t = \text{LSTM}([\textbf{\textit{X}}, \textbf{\textit{V}}_{\!t-1}, \textbf{\textit{a}}^{\!\text{nv}}_{t-1}], \textbf{\textit{h}}^{\!\text{nv}}_{t-1}).
\end{aligned}
\end{equation}
With current navigation state $\textbf{\textit{h}}^{\text{nv}}_t$, the probability
of $k^{th\!}$ navigation action is:
\begin{equation}
\begin{aligned}
p^{\text{nv}}_{t,k}= \text{softmax}_k(\textbf{\textit{v}}^\top_{t,k}\textbf{\textit{W}}^{\text{nv}}~\!\textbf{\textit{h}}^{\!\text{nv}}_t).
\end{aligned}
\label{eq:decision-making}
\end{equation}
Here, $\textbf{\textit{W}}^{\text{nv}\!}$ indicates a learnable parameter matrix. 
The navigation action $\textit{a}^{\text{nv}}_t$ is chosen according to the probability distribution $\{p^{\text{nv}}_{t,k}\}_{k=1}^{K}$.

\noindent \textbf{Basic Agent Implementation.} So far, we have given a brief description of our basic navigation agent from a high-level view, also commonly shared with prior art. In practice, we choose~\cite{tan2019learning} to implement our agent (but not limited to).

\noindent \textbf{Core Idea.} When following instructions, humans do not expect every step to be a ``perfect'' navigation decision, due to current limited visual perception, the inevitable ambiguity in instructions, and the complexity of environments. Instead, when we are uncertain about the future steps, we tend to explore the surrounding first and gather more information to mitigate the ambiguity, and then make a more informed decision. Our core idea is thus to equip an agent with such active exploration/learning ability. To ease understanding, we start with a na\"{i}ve model which is equipped with a simplest exploration function (\S\ref{sec:native}). We then complete the na\"{i}ve model in \S\ref{sec:wte} and \S\ref{sec:mde} and showcase how a learned active exploration policy can greatly improve the navigation performance.
\subsection{A Na\"{i}ve Model with A Simple Exploration Ability}
\label{sec:native}
Here, we consider the most straightforward way of achieving our idea. At each navigation step, the agent simply explores all the navigable views and only one exploration step is allowed for each. This means that the agent explores the first direction, gathers surrounding information and then returns to the original navigation position. Next, it goes one step towards the second navigable direction and turns back. Such one-step exploration process is repeated until all the possible directions have been visited. The information gathered during exploration will be used to support current navigation-decision making.

\begin{figure}[t]
  \centering
      \includegraphics[width=0.99\linewidth]{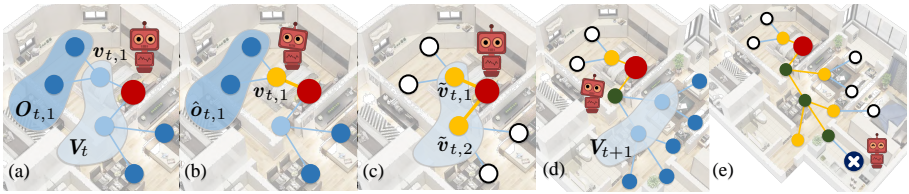}
      \put(-340,28){\fontsize{8pt}{6.2pt}\selectfont $\textbf{\textit{O}}_{t,1}$}
      \put(-310,50){\fontsize{8pt}{6.2pt}\selectfont $\textbf{\textit{v}}_{t,1}$}
      \put(-320,14){\fontsize{8pt}{6.2pt}\selectfont $\textbf{\textit{V}}_{\!\!t}$}
      \put(-273,28){\fontsize{8pt}{6.2pt}\selectfont $\hat{\textbf{\textit{o}}}_{t,1\!}$}
      \put(-249,35){\fontsize{8pt}{6.2pt}\selectfont $\textbf{\textit{v}}_{t,1}$}
      \put(-179,35){\fontsize{8pt}{6.2pt}\selectfont $\tilde{\textbf{\textit{v}}}_{t,1}$}
      \put(-179,15){\fontsize{8pt}{6.2pt}\selectfont $\tilde{\textbf{\textit{v}}}_{t,2}$}
      \put(-122,14){\fontsize{8pt}{6.2pt}\selectfont $\textbf{\textit{V}}_{\!\!t+1}$}
\caption{{$_{\!}$Illustration$_{\!}$ of our$_{\!}$ na\"{i}ve model$_{\!}$ (\S\ref{sec:native}). (a)$_{\!}$ At$_{\!}$ $t^{th\!}$ navigation$_{\!}$ step, the$_{\!}$ agent$_{\!}$ has$_{\!}$ a$_{\!}$  panoramic$_{\!}$ view$_{\!}$ $\!\textbf{\textit{V}}_{\!\!t}$. For $k^{th\!}$ subview, we$_{\!}$ further$_{\!}$ denote$_{\!}$ its$_{\!}$ panoramic$_{\!}$ view$_{\!}$ as$_{\!}$ $\!\textbf{\textit{O}}_{t,k}$. (b) After making a one-step$_{\!}$ exploration$_{\!}$ in$_{\!}$ the$_{\!}$ first$_{\!}$ direction $\textbf{\textit{v}}_{t,1}$, the agent collects information $\hat{\textbf{\textit{o}}}_{t,1\!}$ from $\textbf{\textit{O}}_{t,1}$ via Eq.\!~\ref{equ:ci}. (c) After exploring all the directions, the agent updates his knowledge, \ie, $\{\tilde{\textbf{\textit{v}}}_{t,1},\tilde{\textbf{\textit{v}}}_{t,2}\}$, via Eq.\!~\ref{equ:gtk}. (d) With the updated knowledge,  the agent computes the navigation probability distribution $\{p^{\text{nv}}_{t,k}\}_k$ (Eq.\!~\ref{eq:decision-making2}) and makes a more reliable navigation decision (\ie, $a^{\text{nv}}_{t}\!=\!2$). (e) Visualization of navigation route, where yellow lines are the exploration routes and green circles are navigation landmarks.
}
}
\label{fig:naive}
\end{figure}

Formally, at $t^{th}$ navigation step, the agent has $K$ navigable views, \ie, $\!\textbf{\textit{V}}_{\!t}\!=\!\{\textbf{\textit{v}}_{t,1},\textbf{\textit{v}}_{t,2},\cdots\!,\textbf{\textit{v}}_{t,K}\}$. For $k^{th}$ view, we further denote its $K'^{\!}$ navigable views as $\textbf{\textit{O}}_{t,k}\!=\!\{\textbf{\textit{o}}_{t,k,1},\textbf{\textit{o}}_{t,k,2},\cdots\!, \textbf{\textit{o}}_{t,k,K'}\}$ (see Fig.\!~\ref{fig:naive}(a)). The subscript $(t,k)$ will be omitted for notation simplicity. If the agent makes a one-step exploration in $k^{th}$ direction, he is desired to collect surrounding information from $\textbf{\textit{O}}$. Specifically, keeping current navigation state $\textbf{\textit{h}}^{\!\text{nv}}_t$ in mind, the agent assembles the visual information from $\textbf{\textit{O}}$ by an attention operation (Fig.\!~\ref{fig:naive}(b)):
\begin{equation}
\begin{aligned}
\!\!\!\!\hat{\textbf{\textit{o}}}_{t,k\!}\!=\!\text{att}(\textbf{\textit{O}}, \textbf{\textit{h}}^{\!\text{nv}}_t)\!= \!\sum\nolimits_{k'=1}^{K'}\!\alpha_{k'}\textbf{\textit{o}}_{k'}, ~\!\text{where}~\alpha_{k'\!} = \text{softmax}_{k'}({\textbf{\textit{o}}^\top_{k'\!}}\textbf{\textit{W}}^{\text{att}\!}~\!\textbf{\textit{h}}^{\!\text{nv}}_t).\!\!
\end{aligned}
\label{equ:ci}
\end{equation}
Then, the collected information $\hat{\textbf{\textit{o}}}_{t,k\!}$ is used to update the current visual knowledge $\textbf{\textit{v}}_{t,k}$ about $k^{th}$ view, computed in a residual form (Fig.\!~\ref{fig:naive}(c)):
\begin{equation}
    \tilde{\textbf{\textit{v}}}_{t,k} = \textbf{\textit{v}}_{t,k} + \textbf{\textit{W}}^{\text{o}\!}~\!\hat{\textbf{\textit{o}}}_{t,k}.
\label{equ:gtk}
\end{equation}
In this way, the agent successively makes one-step explorations of all $K$ navi-

\noindent gable views and enriches his corresponding  knowledge. Later, with the updated knowledge  $\{\tilde{\textbf{\textit{v}}}_{t,1},\tilde{\textbf{\textit{v}}}_{t,2},\cdots\!,\tilde{\textbf{\textit{v}}}_{t,K}\}$,  the probability
of making $k^{th}$ navigable action (originated in Eq.\!~\ref{eq:decision-making}) can be formulated as (Fig.\!~\ref{fig:naive}(d)):
\begin{equation}
\begin{aligned}
p^{\text{nv}}_{t,k}= \text{softmax}_k(\tilde{\textbf{\textit{v}}}^\top_{t,k}\textbf{\textit{W}}^{\text{nv}\!}~\!\textbf{\textit{h}}^{\!\text{nv}}_t).
\end{aligned}
\label{eq:decision-making2}
\end{equation}
Through this exploration, the agent should be able to  gather more information from its surroundings, and then make a more reasonable navigation decision. In \S\ref{sec:abs}, we empirically demonstrate that, by equipping the basic agent with such a na\"{i}ve exploration module, we achieve 4$\sim$6\% performance improvement in terms of Successful Rate (SR). This is impressive, as we only allow the agent to make one-step exploration. Another notable issue is that the agent simply explores all the possible directions, resulting in long Trajectory Length (TL)\footnote{Here the routes for navigation and exploration are both involved in TL computation.}. Next we will improve the na\"{i}ve model, by tackling two key issues: ``\textbf{how to decide where to explore}'' (\S\ref{sec:wte}) and ``\textbf{how to make deeper exploration}'' (\S\ref{sec:mde}).

\subsection{Where to Explore}
\label{sec:wte}
In the na\"{i}ve model (\S\ref{sec:native}), the agent conducts exploration of all navigable views at every navigation step. Such a strategy is unwise and brings longer trajectories, and goes against the intuition that exploration is only needed at a few navigation steps, in a few directions. To address this, the agent should learn an \textit{exploration-decision making} strategy, \ie, more actively deciding which direction to explore.

\begin{figure}[t]
      \centering
          \includegraphics[width=0.99\linewidth]{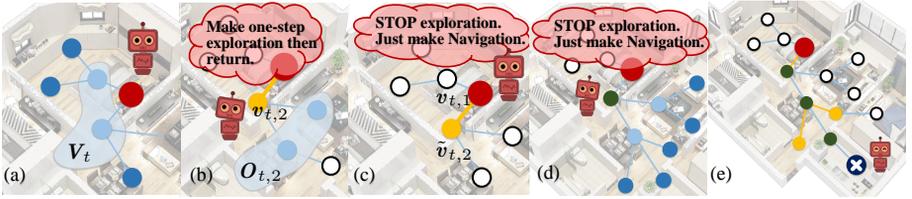}
          \put(-320,14){\fontsize{8pt}{6.2pt}\selectfont $\textbf{\textit{V}}_{\!t}$}
          \put(-249,30){\fontsize{8pt}{6.2pt}\selectfont $\textbf{\textit{v}}_{t,2}$}
          \put(-254,7){\fontsize{8pt}{6.2pt}\selectfont $\textbf{\textit{O}}_{t,2}$}
          \put(-179,35){\fontsize{8pt}{6.2pt}\selectfont $\textbf{\textit{v}}_{t,1}$}
          \put(-179,15){\fontsize{8pt}{6.2pt}\selectfont $\tilde{\textbf{\textit{v}}}_{t,2}$}
    \caption{{$_{\!}$Equip our agent with an exploration-decision making ability\!~(\S\ref{sec:wte}). (a)\!~The agent predicts a probability distribution $\{p^{\text{ep}}_{t,k}\}_{k=1}^{K\!+\!1}$ over exploration action candidates (\ie, Eq.\!~\ref{eq:exd}). (b)\!~According to $\{p^{\text{ep}}_{t,k}\}_{k=1}^{K\!+\!1}$, the most ``valuable'' view is~selected to make a one-step exploration. (c)\!~The agent updates his knowledge $\tilde{\textbf{\textit{v}}}_{t,2}$ and makes a second-round exploration decision (Eq.\!~\ref{eq:exd22}). If \texttt{STOP} action is selected, the agent will make a navigation decision (Eq.\!~\ref{eq:decision-making2}) and start $(t\!+\!1)^{th\!}$ navigation step.
    }
    }
    \label{fig:model2}
\end{figure}

To achieve this, at each navigation step $t$, we let the agent make an exploration decision $a^{\!\text{ep}}_t\!\in\!\{1,\cdots\!,K_{\!}+_{\!}1\}$ from current $K$ navigable views as well as a \texttt{STOP} action. Thus, the exploration action embedding $\textbf{\textit{a}}^{\!\text{ep}}_{t}$ is a vector selected from the visual features of the $K$ navigable views (\ie, $\textbf{\textit{V}}_{\!t}\!=\!\{\textbf{\textit{v}}_{t,1},\textbf{\textit{v}}_{t,2},\cdots\!,\textbf{\textit{v}}_{t,K}\}$), and the \texttt{STOP} action embedding (\ie, $\textbf{\textit{v}}_{t,K\!+\!1}\!=\!\vec{0}$). To learn the exploration-decision making strategy, with current navigation state $\textbf{\textit{h}}^{\!\text{nv}}_t$ and current visual surrounding knowledge  $\textbf{\textit{V}}_{\!t}$, the agent predicts a probability distribution $\{p^{\text{ep}}_{t,k}\}_{k=1}^{K\!+\!1}$ for the $K\!+\!1$ exploration action candidates (Fig.\!~\ref{fig:model2}(a)):
\begin{equation}
\begin{aligned}
\hat{\textbf{\textit{v}}}_{t} &=\text{att}(\!\textbf{\textit{V}}_{\!\!t}, \textbf{\textit{h}}^{\!\text{nv}}_t), ~~~~~~
p^{\text{ep}}_{t,k} &= \text{softmax}_k(\textbf{\textit{v}}^\top_{t,k}\textbf{\textit{W}}^{\text{ep}}~\!\![\hat{\textbf{\textit{v}}}_{t}, \textbf{\textit{h}}^{\!\text{nv}}_t]).
\end{aligned}
\label{eq:exd}
\end{equation}
Then, an exploration action $k^*$ is made according to $\arg\max_{k}p^{\text{ep}}_{t,k}$. If the \texttt{STOP} action is selected (\ie, $k^*\!=\!K\!+\!1$), the agent directly turns to making a navigation decision by Eq.\!~\ref{eq:decision-making}, without exploration. Otherwise, the agent will make a one-step exploration in a most ``valuable'' direction $k^*\!\in\!\{1,\cdots\!,K\}$ (Fig.\!~\ref{fig:model2}(b)). Then, the agent uses the collected information  $\hat{\textbf{\textit{o}}}_{t,k^*}$ (Eq.\!~\ref{equ:ci}) to enrich his
knowledge $\textbf{\textit{v}}_{t,k^*}$ about ${k^{*th}}$ viewpoint (Eq.\!~\ref{equ:gtk}). With the updated knowledge, the agent makes a second-round exploration decision  (Fig.\!~\ref{fig:model2}(c)):
\begin{equation}
\begin{aligned}
\tilde{\textbf{\textit{V}}}_{\!t}\leftarrow\{&\textbf{\textit{v}}_{t,1},\cdots,\tilde{\textbf{\textit{v}}}_{t,k^*},\cdots,\textbf{\textit{v}}_{t,K}\},
~~~~\hat{\textbf{\textit{v}}}_{t} \leftarrow\text{att}(_{\!}\tilde{\textbf{\textit{V}}}_{\!t}, \textbf{\textit{h}}^{\!\text{nv}}_t),\\
&p^{\text{ep}}_{t,k^u} \leftarrow \text{softmax}_{k^u}(\textbf{\textit{v}}^\top_{t,k^u}\textbf{\textit{W}}^{\text{ep}}~\![\hat{\textbf{\textit{v}}}_{t}, \textbf{\textit{h}}^{\!\text{nv}}_t]).
\end{aligned}
\label{eq:exd22}
\end{equation}
Note that the views that have been already explored are removed from the exploration action candidate set, and $k^u$ indicates an exploration action that has not been selected yet. Based on the new exploration probability distribution $\{p^{\text{ep}}_{t,k^u}\}_{k^u}^{K\!+\!1}$, if the \texttt{STOP} action is still not selected, the agent will make a second-round exploration in a new direction. The above multi-round exploration process will be repeated until either the agent is satisfied with his current knowledge about the surroundings (\ie, choosing the \texttt{STOP} decision), or all the $K$ navigable directions are explored. Finally, with the newest knowledge about the surroundings $\tilde{\textbf{\textit{V}}}_{\!t}$, the agent makes a more reasonable navigation decision (Eq.\!~\ref{eq:decision-making2}, Fig.\!~\ref{fig:model2}(d)). Our experiments in \S\ref{sec:abs} show that, when allowing the agent to actively select navigation directions, compared with the na\"{i}ve model, TL is greatly decreased and even SR is improved (as the agent focuses on the most valuable directions).

\subsection{Deeper Exploration}
\label{sec:mde}
\begin{figure}[t]
      \centering
          \includegraphics[width=0.99\linewidth]{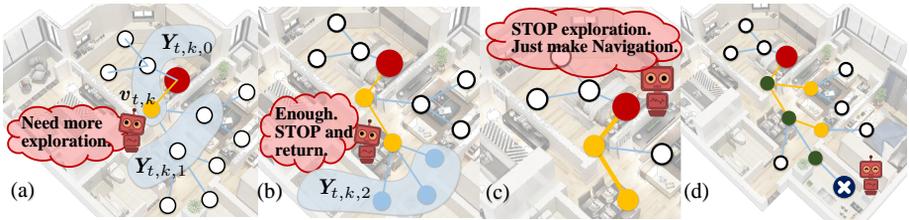}
          \put(-285,64){\fontsize{8pt}{6.2pt}\selectfont $\textbf{\textit{Y}}_{\!\!t,k,0}$}
          \put(-300,45){\fontsize{8pt}{6.2pt}\selectfont $\textbf{\textit{v}}_{t,k}$}
          \put(-295,18){\fontsize{8pt}{6.2pt}\selectfont $\textbf{\textit{Y}}_{\!\!t,k,1}$}
          \put(-225,10){\fontsize{8pt}{6.2pt}\selectfont $\textbf{\textit{Y}}_{\!\!t,k,2}$}

    \caption{{Our full model can actively make multi-direction, multi-step exploration. (a) The agent is in 1$^{st\!}$ exploration step ($s\!=\!1$), starting from $k^{th\!}$ view at $t^{th\!}$ navigation step. According to the exploration probability $\{p^{\text{ep}}_{s,k'\!}\}_{k'\!}$ (Eq.\!~\ref{eq:decision11}), the agent decides to make a further step exploration. (b) At 2$^{nd\!}$ exploration step, the agent decides to finish the exploration of $k^{th\!}$ view. (c) The agent thinks there is no other direction worth exploring, then makes a navigation decision based on the updated knowledge.
    }
    }
    \label{fig:model3}
\end{figure}

So far, our agent is able to make explorations only when necessary. Now we focus on how to let him conduct multi-step exploration, instead of simply constraining the maximum exploration length as one.  Ideally, during the exploration of a certain direction, the agent should be able to go ahead a few steps until sufficient information is collected. To model such a sequential exploration decision-making process, we design a recurrent network based exploration module, which also well generalizes to the cases discussed in \S\ref{sec:native} and \S\ref{sec:wte}. Specifically, let us assume that the agent starts an exploration episode from $k^{th\!}$ view $\textbf{\textit{v}}_{t,k\!}$ at $t^{th\!}$ navigation step (Fig.\!~\ref{fig:model3}(a)). At an exploration step $s$, the agent perceives the surroundings with a panoramic view and collects information from $K'$ navigable views $\textbf{\textit{Y}}_{\!\!t,k,s\!}\!=_{\!}\!\{\textbf{\textit{y}}_{t,k,s,1},\textbf{\textit{y}}_{t,k,s,2},\cdots\!,\textbf{\textit{y}}_{t,k,s,K'}\}$. With such a definition, we have $\!\textbf{\textit{Y}}_{\!\!t,k,0\!}\!=\!\!\textbf{\textit{V}}_{\!\!t}$. In \S\ref{sec:native} and \S\ref{sec:wte}, for $k^{th\!}$ view at $t^{th\!}$ navigation step, its panoramic view $\textbf{\textit{O}}_{t,k}$ is also $\textbf{\textit{Y}}_{\!\!t,k,1}$. The subscript $(t,k)$ will be omitted for notation simplicity.

\noindent\textbf{Knowledge Collection During Exploration:} As the exploration module is in a recurrent form, the agent has a specific state $\textbf{\textit{h}}^{\!\text{ep}}_s$  at $s^{th}$ exploration step. With $\textbf{\textit{h}}^{\!\text{ep}}_s$, the agent actively collects knowledge by assembling the surrounding information $\textbf{\textit{Y}}_{\!s}$ using an attention operation (similar to Eq.\!~\ref{equ:ci}):
\begin{equation}
\begin{aligned}
\hat{\textbf{\textit{y}}}_{s} = \text{att}(\textbf{\textit{Y}}_{\!\!s}, \textbf{\textit{h}}^{\!\text{ep}}_s).
\end{aligned}
\label{eq:infogather}
\end{equation}

\noindent\textbf{Knowledge Storage During Exploration:} As the agent performs multi-step exploration, the learned knowledge $\hat{\textbf{\textit{y}}}_{s}$ is stored in a memory network:
\begin{equation}
\begin{aligned}
\textbf{\textit{h}}^{\!\text{kw}}_{s} = \text{LSTM}^{\text{kw}}(\hat{\textbf{\textit{y}}}_{s}, \textbf{\textit{h}}^{\!\text{kw}}_{s-1}),
\end{aligned}
\label{eq:infostore}
\end{equation}
which will eventually be used for supporting navigation-decision making.

\noindent\textbf{Sequential Exploration-Decision Making for Multi-Step Exploration:}  Next, the agent needs to decide whether or not to choose a new direction for further exploration. In the exploration action space, the agent either selects one direction from the  current $K'^{\!}$ reachable views to explore or stops the current exploration episode and returns to the original position at $t^{th\!}$ navigation step. The exploration action $a^{\text{ep\!}}_s$ is represented as a vector $\textbf{\textit{a}}^{\text{ep\!}}_s$ from the visual features of the $K'^{\!}$ navigable views (\ie, $\textbf{\textit{Y}}_{\!s\!}\!=\!\{\textbf{\textit{y}}_{s,1},\textbf{\textit{y}}_{s,2},\cdots\!,\textbf{\textit{y}}_{s,K'}\}$), as well as the \texttt{STOP} action embedding (\ie, $\textbf{\textit{y}}_{s,K'\!+\!1\!}\!=\!\vec{0}$). $a^{\text{ep}\!}_s$  is predicted according to the current exploration state $\textbf{\textit{h}}^{\text{ep}\!}_s$ and collected information $\textbf{\textit{h}}^{\text{kw}\!}_{s}$. Hence, the computation of $\textbf{\textit{h}}^{\text{ep}\!}_s$ is conditioned on the current navigation state $\textbf{\textit{h}}^{\text{nv}\!}_t$,  history exploration views $\{\textbf{\textit{Y}}_{\!1},\textbf{\textit{Y}}_{\!2},\cdots\!,\textbf{\textit{Y}}_{\!s-1}\}$, and previous exploration actions $\{\textbf{\textit{a}}^{\text{ep}}_1,\textbf{\textit{a}}^{\text{ep}}_2,\cdots\!,\textbf{\textit{a}}^{\text{ep}}_{s-1}\}$:
\begin{equation}
\begin{aligned}
\textbf{\textit{h}}^{\!\text{ep}}_s = \text{LSTM}^{\text{ep}}([\textbf{\textit{h}}^{\!\text{nv}}_t, \textbf{\textit{Y}}_{\!\!s-1}, \textbf{\textit{a}}^{\text{ep}}_{s-1}], \textbf{\textit{h}}^{\!\text{ep}}_{s-1}), ~\text{where}~\textbf{\textit{h}}^{\!\text{ep}}_0\!=\!\textbf{\textit{h}}^{\!\text{nv}}_t.
\end{aligned}
\end{equation}
For $k'^{th}$ exploration action candidate (reachable view), its probability is:
\begin{equation}
\begin{aligned}
p^{\text{ep}}_{s,k'}= \text{softmax}_{k'}(\textbf{\textit{y}}^\top_{s,k'}\textbf{\textit{W}}^{\text{ep}}~\![\textbf{\textit{h}}^{\!\text{kw}}_{s},\textbf{\textit{h}}^{\!\text{ep}}_s]).
\end{aligned}
\label{eq:decision11}
\end{equation}
The exploration action $\textit{a}^{\text{ep}}_s$ is chosen according to $\{p^{\text{ep}}_{s,k'}\}_{k'=1}^{K'\!+\!1}$.

\noindent\textbf{Multi-Round Exploration-Decision Making for Multi-Direction Exploration:} After $S$-step exploration, the agent chooses the \texttt{STOP} action when he thinks sufficient information along a certain direction $k$ has been gathered (Fig.~\ref{fig:model3} (b)). He goes back to the start point at $t^{th\!}$ navigation step and updates his knowledge about $k^{th\!}$ direction, \ie, $\textbf{\textit{v}}_{t,k}$, with the gathered information $\textbf{\textit{h}}^{\!\text{kw}}_{S}$. Thus, Eq.\!~\ref{equ:gtk} is improved as:
\begin{equation}
 \tilde{\textbf{\textit{v}}}_{t,k} =\textbf{\textit{v}}_{t,k} +
 \textbf{\textit{W}}^{\text{o}}~\!\textbf{\textit{h}}^{\!\text{kw}}_{S}.
\label{equ:gtk3}
\end{equation}
With the updated knowledge regarding the surroundings, the agent makes a second-round exploration decision:
\begin{equation}
\begin{aligned}
\tilde{\textbf{\textit{V}}}_{\!t}\leftarrow\{&\textbf{\textit{v}}_{t,1},\cdots,\tilde{\textbf{\textit{v}}}_{t,k},\cdots,\textbf{\textit{v}}_{t,K}\},
~~~~~~\hat{\textbf{\textit{v}}}_{t} \leftarrow\text{att}(\tilde{\textbf{\textit{V}}}_{\!t}, \textbf{\textit{h}}^{\!\text{nv}}_t),\\
&p^{\text{ep}}_{t,k^u} \leftarrow \text{softmax}_{k^u}(\textbf{\textit{v}}^\top_{t,k^u}\textbf{\textit{W}}^{\text{ep}}~\![\hat{\textbf{\textit{v}}}_{t}, \textbf{\textit{h}}^{\!\text{nv}}_t]).
\end{aligned}
\end{equation}
Again, $k^{u\!}$ indicates an exploration action that has not been selected yet. Then the agent can make another round of exploration in a new direction, until he chooses the \texttt{STOP} action (\ie, the collected information is enough to help make a confident navigation decision), or has explored all $K$ directions (Fig.\!~\ref{fig:model3}(c)).

\noindent\textbf{Exploration-Assisted Navigation-Decision Making:} After multi-round multi-step exploration, with the newest knowledge $\!\tilde{\textbf{\textit{V}}}_{\!t}$ about the surroundings, the agent makes a more reliable navigation decision (Eq.~\!\ref{eq:decision-making2}): $p^{\text{nv}}_{t,k}\!=\!\text{softmax}_k(\tilde{\textbf{\textit{v}}}^\top_{t,k\!}\textbf{\textit{W}}^{\text{nv}\!}\textbf{\textit{h}}^{\text{nv}}_t)$. Then, at $(t\!+\!1)^{th}$ navigation step, the agent makes multi-step explorations in several directions (or can even omit exploration) and then chooses a new navigation action. In \S\ref{sec:abs}, we will empirically demonstrate that our full model gains the highest SR score with only slightly increased TL.

\noindent\textbf{Memory based Late Action-Taking Strategy:} After finishing exploration towards a certain direction, if directly ``going back'' the start position and making next-round exploration/navigation, it may cause a lot of revisits. To alleviate this, we let the agent store the visited views during exploration in an outside memory. The agent then follows a late action-taking strategy, \ie, moving only when it is necessary. When the agent decides to stop his exploration at a direction, he stays at his current position and ``images'' the execution of his following actions without really going back. When he needs to visit a new point that is not stored in the memory, he will go to that point directly and updates the memory accordingly. Then, again, holding the position until he needs to visit a new point that is not met before. Please refer to the supplementary for more details.

\subsection{Training}
Our entire agent model is trained with two distinct learning paradigms, \ie, 1) imitation learning, and 2) reinforcement learning.

\noindent \textbf{Imitation Learning (IL).} In IL, an agent is forced to mimic the behavior of its teacher. Such a strategy has been proved effective in VLN\!~\cite{mei2016listen,anderson2018vision,wang2018look,fried2018speaker,tan2019learning,ma2019regretful,wang2019reinforced}. Specifically, at navigation step $t$, the teacher provides the teacher action $a_t^*\!\in\!\{1,\cdots,K\}$, which selects the next navigable viewpoint on the shortest route from the current viewpoint to the target viewpoint. The negative log-likelihood of the demonstrated action is computed as the IL loss:
\begin{equation}
    \mathcal{L}_{\text{IL}}^{\text{nv}}=\sum\nolimits_{t}-\log~p^{\text{nv}}_{t,a^{*}_t}.
\label{equ:il}
\end{equation}
The IL loss for the exploration is defined as:
\begin{equation}
    \mathcal{L}_{\text{IL}}^{\text{ep}}=\sum\nolimits_{t}\sum\nolimits^S_{s=0}-\log~p^{\text{ep}}_{s,a^{*}_{t+s}},
\label{equ:il2}
\end{equation}
where $S$ is the maximum number of steps allowed for exploration. At $t^{th\!}$ navigation step, the agent performs $S$-step exploration, simply imitating the teacher's navigation actions from $t$ to $t\!+\!S$ steps. Though the goals of navigation and exploration are different, here we simply use the teacher navigation actions to guide the learning of exploration, which helps the exploration module learn better representations, and quickly obtain an initial exploration policy.


\noindent \textbf{Reinforcement Learning (RL).} Through IL, the agent can learn an off-policy that
works relatively well on seen scenes, but it is biased towards copying the route introduced by the teacher, rather than learning how to recover from its erroneous behavior in an unseen environment\!~\cite{wang2018look}. Recent methods\!~\cite{zhu2017target,wang2018look,tan2019learning,wang2019reinforced} demonstrate that the on-policy RL method Advantage Actor-Critic (A2C)\!~\cite{mnih2016asynchronous} can help the agent explore the state-action space outside the demonstration path.

For RL based navigation learning, our agent samples a navigation action from the
distribution $\{p^{\text{nv}}_{t,k}\}_{k=1\!}^{K}$ (see Eq.\!~\ref{eq:decision-making}) and learns from rewards. Let us denote the reward after taking a navigation action $a^{\text{nv}}_t$  at current view $v_t$ as $r(v_t, a^{\text{nv}}_t)$. As in\!~\cite{tan2019learning,wang2019reinforced}, at each non-stop step $t$, $r^{\text{nv}}(v_t, a^{\text{nv}}_t)$ is
the change in the distance to the target navigation location. At the final step $T$, if the agent stops within 3 meters of  the target location, we set $r^{\text{nv}}(v_T, a^{\text{nv}}_T)\!=\!+3$; otherwise $r^{\text{nv}}(v_T, a^{\text{nv}}_T)\!=\!-3$. Then, to incorporate the influence of the action $a^{\text{nv}}_t$ on the future and account for the local greedy search, the total accumulated return with a discount factor is adopted: $R^{\text{nv}}_t\!=\!\sum\nolimits_{t'=t}^{T}\gamma^{t'-t}r^{\text{nv}}(v_{t'}, a^{\text{nv}}_{t'})$,
where the discounted factor $\gamma$ is set as 0.9. In A2C, our agent can be viewed as an actor and a state-value function  $b^{\text{nv}}(\textbf{\textit{h}}_t)$, viewed as critic, is evaluated. For training, the actor aims to minimize the negative log-probability of action $a^{\text{nv}}_t$ scaled by $R^{\text{nv}}_t\!-\!b^{\text{nv}}(\textbf{\textit{h}}^{\!\text{nv}}_t)$ (known as the \textit{advantage} of action $a^{\text{nv}}_t$), and the critic $b^{\text{nv}}(\textbf{\textit{h}}_t)$ aims to minimize the Mean-Square-Error between $R^{\text{nv}}_t$ and the estimated value:
\begin{equation}
    \mathcal{L}_{\text{RL}}^{\text{nv}}=-\sum\nolimits_t(R^{\text{nv}}_t-b^{\text{nv}}(\textbf{\textit{h}}^{\!\text{nv}}_t))\text{log~}p^{\text{nv}}_{t,a^{\text{nv}}_t} + \sum\nolimits_t(R^{\text{nv}}_t-b^{\text{nv}}(\textbf{\textit{h}}^{\text{nv}}_t))^2.
 \label{equ:rl1}
\end{equation}

For RL-based exploration learning, we also adopt on-policy A2C for training. Specifically, let us assume a set of explorations $\{a^{\text{ep}}_{t,k,s}\}^{S_{t,k}}_{s=1}$ are made in a certain direction $k$ at navigation step $t$, and the original navigation action (before exploration) is $a'^{\text{nv}}_t$. Also assume that the exploration-assisted navigation action (after exploration) is $a^{\text{nv}}_t$. The basic reward $r^{\text{ep}}(v_t,\{a^{\text{ep}}_{t,k,s}\}_s)$ for the exploration actions $\{a^{\text{ep}}_{t,k,s}\}_s$ is defined as:
\begin{equation}
r^{\text{ep}}(v_t,\{a^{\text{ep}}_{t,k,s}\}_s) = r^{\text{nv}}(v_t, a^{\text{nv}}_t)-r^{\text{nv}}(v_t,a'^{\text{nv}}_t).
\end{equation}
This means that, after making explorations $\{a^{\text{ep}}_{t,k,s}\}_s$ at $t^{th}$ navigation step in $k^{th\!}$ direction, if the new navigation decision $a^{\text{nv}}_t$ is better than the original one $a'^{\text{nv}}_t$, \ie, helps the agent make a better navigation decision, a positive exploration reward will be assigned. More intuitively, such an exploration reward represents the benefit that this set of explorations $\{a^{\text{ep}}_{t,k,s}\}_s$ could bring for the navigation. We average $r^{\text{ep}}(v_t,\{a^{\text{ep}}_{t,k,s}\}_s)$ to each exploration action $a^{\text{ep}}_{t,k,s}$ as the immediate reward, \ie, $r^{\text{ep}}(v_t,a^{\text{ep}}_{t,k,s})\!=\!\frac{1}{S_{t,k}}r^{\text{ep}}(v_t,\{a^{\text{ep}}_{t,k,s}\}_s)$. 
In addition, to limit the length of exploration, we add a negative term $\beta$ (=$-$0.1) to the reward of each exploration step. Then, the total accumulated discount return for an exploration action $a^{\text{ep}}_{t,k,s}$ is defined as: $R^{\text{ep}}_{t,k,s}\!=\!\sum\nolimits_{s'=s}^{S_{t,k}}\gamma^{s'-s}(r^{\text{ep}}(v_t,a^{\text{ep}}_{t,k,s'})\!+\!\beta)$. The RL loss for the exploration action $a^{\text{ep}}_{t,k,s}$ is defined as:
\begin{equation}
\mathcal{L}(a^{\text{ep}}_{t,k,s})=-(R^{\text{ep}}_{t,k,s}-b^{\text{ep}}(\textbf{\textit{h}}^{\!\text{ep}}_{t,k,s}))\text{log~}p^{\text{ep}}_{t,k,a^{\text{ep}}_{t,k,s}}\!+
    (R^{\text{ep}}_{t,k,s}-b^{\text{ep}}(\textbf{\textit{h}}^{\!\text{ep}}_{t,k,s}))^2,
\end{equation}
where $b^{\text{ep}}$ is the critic. Then, similar to Eq.~\ref{equ:rl1}, the
RL loss for all the exploration actions is defined as:
\begin{equation}
    \mathcal{L}_{\text{RL}}^{\text{ep}}=-\sum\nolimits_t\sum\nolimits_k\sum\nolimits_s\mathcal{L}(a^{\text{ep}}_{t,k,s}).
 \label{equ:rl2}
\end{equation}

\noindent \textbf{Curriculum Learning for Multi-Step Exploration.}   
During training, we find that, once the exploration policy is updated, the model easily suffers from extreme variations in gathered information,  particularly for long-term exploration, making the training jitter. To avoid this, we adopt curriculum learning\!~\cite{bengio2009curriculum} to train our agent with incrementally improved exploration length. Specifically, in the beginning, the maximum exploration length is set to 1. After the training loss converges, we use current parameters to initialize the training of the agent with at most 2-step exploration. In this way, we train an agent with at most 6-step exploration (due to the limited GPU memory and time). This strategy greatly improves the convergence speed (about $\times$8 faster) with no noticeable diminishment in performance. Experiments related to the influence of the maximum exploration length can be found in \S\ref{sec:abs}.

\noindent \textbf{Back Translation Based Training Data Augmentation.} Following\!~\cite{fried2018speaker,tan2019learning}, we use back translation to augment training data. The basic idea is that, in addition to training a \textit{navigator} that finds  the correct route in an environment according to the given instructions, an auxiliary \textit{speaker} is trained for generating an instruction given a route inside an environment. In this way, we generate extra instructions for $176$k unlabeled routes in Room-to-Room\!~\cite{anderson2018vision} training environments. After training the agent on the labeled samples from the Room-to-Room training set, we use the back translation augmented data for fine-tuning.

\section{Experiment}
\label{sec:exp}
\subsection{Experimental Setup}
\noindent\textbf{Dataset.} We conduct experiments on the Room-to-Room (R2R) dataset\!~\cite{anderson2018vision}, which has $10,800$ panoramic views in $90$ housing environments, and $7,189$ paths sampled from its navigation graphs. Each path is associated with three ground-truth navigation instructions. R2R is split into four sets: \texttt{training}, \texttt{validation} \texttt{seen}, \texttt{validation} \texttt{unseen}, and \texttt{test} \texttt{unseen}. There are no overlapping environments between the unseen and training sets.


\noindent\textbf{Evaluation Metric.} As in conventions\!~\cite{anderson2018vision,fried2018speaker}, five metrics are used for evaluation: \textit{Success Rate} (SR), \textit{Navigation Error} (NE), \textit{Trajectory Length} (TL), \textit{Oracle success Rate} (OR), and \textit{Success rate weighted by Path Length} (SPL).


\noindent\textbf{Implementation$_{\!}$ Detail.} As$_{\!}$ in\!~\cite{fried2018speaker,anderson2018vision,wang2019reinforced,wang2018look},
the$_{\!}$ viewpoint$_{\!}$ embedding$_{\!}$ $\textbf{\textit{v}}_{t,k\!}$ is$_{\!}$ a$_{\!}$ concatenation$_{\!}$ of$_{\!}$ image$_{\!}$ feature$_{\!}$ (from$_{\!}$ an$_{\!}$ ImageNet\!~\cite{ILSVRC15}$_{\!}$ pre-trained$_{\!}$ ResNet-152\!~\cite{he2016deep}) and$_{\!}$ a$_{\!}$ 4-$d$$_{\!}$ orientation$_{\!}$ descriptor.  A$_{\!}$ bottleneck$_{\!}$ layer$_{\!}$ is$_{\!}$ applied$_{\!}$ to$_{\!}$ reduce$_{\!}$ the$_{\!}$ dimension$_{\!}$ of$_{\!}$ $\textbf{\textit{v}}_{t,k\!}$  to$_{\!}$ 512$_{\!}$. Instruction$_{\!}$ embeddings$_{\!}$ $\textbf{\textit{X}}_{\!}$ are$_{\!}$ obtained$_{\!}$ from$_{\!}$ an$_{\!}$ LSTM$_{\!}$ with$_{\!}$ a 512$_{\!}$ hidden$_{\!}$ size$_{\!}$. For$_{\!}$ each$_{\!}$ LSTM$_{\!}$ in$_{\!}$ our$_{\!}$ exploration$_{\!}$ module, the$_{\!}$ hidden$_{\!}$ size$_{\!}$ is$_{\!}$ 512.  For$_{\!}$ back$_{\!}$ translation, the$_{\!}$ speaker$_{\!}$ is$_{\!}$ implemented$_{\!}$ as$_{\!}$  described$_{\!}$ in\!~\cite{fried2018speaker}.


\subsection{Comparison Results}
\label{sec:cr}


\begin{table}[t]
\begin{center}
	\caption{{Comparison results on \texttt{validation} \texttt{seen}, \texttt{validation} \texttt{unseen}, and \texttt{test} \texttt{unseen} sets of R2R~\cite{anderson2018vision} under \textbf{Single Run} setting (\S\ref{sec:cr}). For compliance with the evaluation server, we report SR as fractions.  $^*$: back translation augmentation.}}
\label{table:leadersingle}
        \resizebox{1\textwidth}{!}{
		\setlength\tabcolsep{1pt}
		\renewcommand\arraystretch{1.0}
\begin{tabular}{c||ccccc|ccccc|ccccc}
\hline \thickhline
\rowcolor{mygray}
~ &  \multicolumn{15}{c}{Single Run Setting} \\
\cline{2-16}
\rowcolor{mygray}
~ &  \multicolumn{5}{c|}{\texttt{validation} \texttt{seen}} & \multicolumn{5}{c|}{\texttt{validation} \texttt{unseen}} & \multicolumn{5}{c}{\texttt{test} \texttt{unseen}} \\
\cline{2-16}
\rowcolor{mygray}
\multirow{-3}{*}{Models} &\textbf{SR}$\uparrow$ &NE$\downarrow$ &TL$\downarrow$ &OR$\uparrow$  &SPL$\uparrow$ &\textbf{SR}$\uparrow$ &NE$\downarrow$ &TL$\downarrow$ &OR$\uparrow$  &SPL$\uparrow$ &\textbf{SR}$\uparrow$ &NE$\downarrow$ &TL$\downarrow$ &OR$\uparrow$ &SPL$\uparrow$\\
\hline
\hline
Random  & 0.16 & 9.45 & 9.58 & 0.21 & - & 0.16 & 9.23 & 9.77 & 0.22  & - & 0.13 & 9.77 & 9.93 & 0.18  & 0.12\\
Student-Forcing~\cite{anderson2018vision} & 0.39 & 6.01 & 11.3 & 0.53  & - & 0.22 & 7.81 & 8.39 & 0.28  & - & 0.20 & 7.85 & 8.13 & 0.27  & 0.18\\
RPA~\cite{wang2018look} & 0.43 & 5.56 & 8.46 & 0.53  & - & 0.25 & 7.65 & 7.22  & 0.32  & - & 0.25 & 7.53 & 9.15 & 0.33 & 0.23 \\
E-Dropout~\cite{tan2019learning} & 0.55 & 4.71 & 10.1 & - & 0.53 & 0.47 & 5.49 & 9.37 & - & 0.43 & - & - & - & -  & - \\
Regretful~\cite{ma2019regretful} & 0.65 & 3.69 & - & 0.72 & 0.59 & 0.48 & 5.36 & - & 0.61 & 0.37 & - & - & - & - & -\\
\textbf{Ours} & 0.66 & 3.35 & 19.8 & 0.79 & 0.49 & 0.55 & 4.40 & 19.9 & 0.70 & 0.38 & 0.56 & 4.77 & 21.0 & 0.73 & 0.37\\
\hline
Speaker-Follower~\cite{fried2018speaker}* & 0.66 & 3.36 & - & 0.74 & - & 0.36 & 6.62 & - & 0.45  & - & 0.35 & 6.62 & 14.8 & 0.44 & 0.28\\
RCM~\cite{wang2019reinforced}*& 0.67 & 3.53 & 10.7 & 0.75 & - & 0.43 & 6.09 & 11.5 & 0.50 & - & 0.43 & 6.12 & 12.0 & 0.50 & 0.38 \\
Self-Monitoring~\cite{ma2019self}*& 0.67 & 3.22 & - & 0.78 & 0.58 & 0.45 & 5.52 & - & 0.56 & 0.32 & 0.43 & 5.99 & 18.0 & 0.55 & 0.32 \\
Regretful~\cite{ma2019regretful}* & 0.69 & 3.23 & - & 0.77  & 0.63 & 0.50 & 5.32 & - & 0.59  & 0.41 & 0.48 & 5.69 & 13.7 & 0.56  & 0.40 \\
E-Dropout~\cite{tan2019learning}* &0.62 & 3.99 & 11.0 & - & 0.59 & 0.52 & 5.22 & 10.7 & - &0.48 & 0.51 & 5.23 & 11.7 & 0.59  & 0.47 \\
Tactical Rewind~\cite{ke2019tactical}*& - & - & - & - & -&0.56 & 4.97  & 21.2 & - &0.43 & 0.54 & 5.14 & 22.1 & 0.64 & 0.41 \\
AuxRN~\cite{zhu2019vision}*&0.70 &3.33 & - & 0.78 &0.67&0.55 &5.28 & - &0.62 & 0.50& 0.55 &5.15 & - &0.62 & 0.51 \\
\textbf{Ours}*& \textbf{0.70} & 3.20 & 19.7 & 0.80 & 0.52 & \textbf{0.58} & 4.36 & 20.6 & 0.70 & 0.40  & \textbf{0.60} & 4.33 & 21.6 & 0.71  & 0.41 \\
\hline
\end{tabular}
}
\end{center}
\end{table}

\begin{table}[t]
\begin{center}
	\caption{{Comparison results on \texttt{test} \texttt{unseen} set of R2R~\cite{anderson2018vision}, under \textbf{Pre-Explore} and \textbf{Beam Search}  settings (\S\ref{sec:cr}). To comply with the evaluation server, we report SR as fractions.  $^*$: back translation augmentation. $-$: unavailable statistics. $\dagger$: a different beam search strategy is used, making the scores uncomparable.}}
\label{table:leader}
        \resizebox{0.85\textwidth}{!}{
		\setlength\tabcolsep{5pt}
		\renewcommand\arraystretch{1.0}
\begin{tabular}{c||ccccc||ccc}
\hline \thickhline
\rowcolor{mygray}
~ &  \multicolumn{5}{c||}{Pre-Explore Setting} & \multicolumn{3}{c}{Beam Search Setting} \\
\cline{2-9}
\rowcolor{mygray}
~ &  \multicolumn{8}{c}{\texttt{test} \texttt{unseen}} \\
\cline{2-9}
\rowcolor{mygray}
\multirow{-3}{*}{Models} &\textbf{SR}$\uparrow$ &NE$\downarrow$ &TL$\downarrow$ &OR$\uparrow$  &SPL$\uparrow$ &\textbf{SR}$\uparrow$ &TL$\downarrow$ &SPL$\uparrow$ \\
\hline
\hline
Speaker-Follower~\cite{fried2018speaker}* & - & - & - & - & - &0.53 &1257.4  &0.01\\
RCM~\cite{wang2019reinforced}*&0.60 &4.21 &9.48 &0.67  &0.59 &0.63 &357.6  &0.02\\
Self-Monitoring~\cite{ma2019self}*& - & - & - & - & - &0.61 &373.1  &0.02\\
E-Dropout~\cite{tan2019learning}* &0.64 &3.97 &9.79 &0.70 &0.61 &0.69 &686.8  &0.01\\
AuxRN~\cite{zhu2019vision}*&0.68 &3.69 & - &0.75 &0.65& 0.70 & $\dagger$ & $\dagger$ \\
\textbf{Ours}*&\textbf{0.70} &3.30 &9.85 &0.77 & 0.68 & \textbf{0.71} &176.2 &0.05\\
\hline
\end{tabular}
}
\end{center}
\end{table}

\noindent\textbf{Performance Comparisons Under Different VLN Settings.} We extensively evaluate our performance under three different VLN setups in R2R.

\noindent\textit{(1) Single Run Setting:} This is the basic setup in R2R, where the agent conducts navigation by selecting the actions in a step-by-step, greedy manner. The agent is not allowed to: 1) run multiple trials, 2) explore or map the test environments before starting. Table\!~\ref{table:leadersingle} reports the comparison results under such a setting. The following are some essential observations. \textbf{i)} Our agent outperforms other competitors on the main metric SR, and some other criteria, \eg, NE and OR. For example, in terms of SR, our model improves AuxRN~\cite{zhu2019vision} 3\% and 5\%, on \texttt{validation} \texttt{unseen} and \texttt{test} \texttt{unseen} sets, respectively, demonstrating our strong generalizability.  \textbf{ii)} Our agent without data augmentation already outperforms many existing methods on SR and NE.  \textbf{iii)} Our TL and SPL scores are on par with current art, with considering exploration routes into the metric computation. \textbf{iv)} If considering the routes for pure navigation, on \texttt{validation} \texttt{unseen} set, our TL is only about 9.4.

\noindent\textit{(2) Pre-Explore Setting:} This setup, first introduced by\!~\cite{wang2019reinforced}, allows the agent to pre-explore the unseen environment before conducting navigation. In\!~\cite{wang2019reinforced}, the agent learns to adapt to the unseen environment through semi-supervised methods, using only pre-given instructions\!~\cite{wang2019reinforced}, without paired routes. Here, we follow a more strict setting, as in\!~\cite{tan2019learning}, where only the unseen environments can be accessed. Specifically, we use back translation to synthesize instructions for routes \textit{sampled} from the unseen environments and fine-tune the agent on the synthetic data. As can be seen from Table\!~\ref{table:leader}, the performance of our method is significantly better than current state-of-the-art methods\!~\cite{wang2019reinforced,tan2019learning,zhu2019vision}.

\noindent\textit{(3) Beam Search Setting:} Beam search was originally used in\!~\cite{fried2018speaker} to optimize SR
 metric. Given an instruction, the agent is allowed to collect multiple candidate routes to score and pick the best one\!~\cite{ke2019tactical}. Following\!~\cite{fried2018speaker,tan2019learning}, we use the speaker to estimate the candidate routes and pick the best one as the final result. As shown in Table\!~\ref{table:leader}, our performance is better than previous methods.

\begin{table}[t]
\begin{center}
	\caption{{Ablation study on the \texttt{validation} \texttt{seen} and \texttt{validation} \texttt{unseen} sets of R2R~\cite{anderson2018vision} under the \textbf{Single Run} setting. See \S\ref{sec:abs} for details. }}
\label{table:abs}
        \resizebox{1\textwidth}{!}{
		\setlength\tabcolsep{1pt}
		\renewcommand\arraystretch{1.0}
\begin{tabular}{c|c||ccccc|ccccc}
\hline \thickhline
\rowcolor{mygray}
& & \multicolumn{10}{c}{Single Run Setting} \\
\cline{3-12}
\rowcolor{mygray}
& & \multicolumn{5}{c|}{\texttt{validation} \texttt{seen}} & \multicolumn{5}{c}{\texttt{validation} \texttt{unseen}} \\
\cline{3-12}
\rowcolor{mygray}
\multicolumn{1}{c|}{\multirow{-3}{*}{Aspect}} &\multirow{-3}{*}{Model} &\textbf{SR}$\uparrow$ &NE$\downarrow$ &TL$\downarrow$ &OR$\uparrow$ &SPL$\uparrow$ &\textbf{SR}$\uparrow$ &NE$\downarrow$ &TL$\downarrow$ &OR$\uparrow$  &SPL$\uparrow$\\
\hline
\hline
Basic agent &\textit{w/o.} any exploration &0.62 & 3.99 & 11.0 & 0.71 & 0.59 & 0.52 & 5.22 & 10.7 & 0.58 &0.48 \\
\hline
&Our na\"{i}ve model (\S\ref{sec:native})&\multirow{2}{*}{0.66} &\multirow{2}{*}{3.55} &\multirow{2}{*}{40.9} &\multirow{2}{*}{0.81} &\multirow{2}{*}{0.19} &\multirow{2}{*}{0.54} &\multirow{2}{*}{4.76} &\multirow{2}{*}{35.7} &\multirow{2}{*}{0.71} &\multirow{2}{*}{0.16} \\
\specialrule{0em}{-0.5pt}{-1.5pt}
&{\color{ggray}\textit{1-step exploration+all directions}} &&&&&&&&&&\\
&\textit{w}. exploration decision (\S\ref{sec:wte})&\multirow{2}{*}{0.66} &\multirow{2}{*}{3.72} &\multirow{2}{*}{12.2} &\multirow{2}{*}{0.76} &\multirow{2}{*}{0.53} &\multirow{2}{*}{0.55} &\multirow{2}{*}{4.82} &\multirow{2}{*}{13.7} &\multirow{2}{*}{0.66} &\multirow{2}{*}{0.42}\\
\specialrule{0em}{-0.5pt}{-1.5pt}
&{\color{ggray}\textit{1-step exploration+parts of directions}}&&&&&&&&&&\\
&\textit{w}. further exploration  &\multirow{2}{*}{0.70} &\multirow{2}{*}{3.15} &\multirow{2}{*}{69.6} &\multirow{2}{*}{0.95} &\multirow{2}{*}{0.13} &\multirow{2}{*}{0.60} &\multirow{2}{*}{4.27} &\multirow{2}{*}{58.8} &\multirow{2}{*}{0.89} &\multirow{2}{*}{0.12}\\
\specialrule{0em}{-0.5pt}{-1.5pt}
\multirow{-6}{*}{Component}&{\color{ggray}\textit{at most 4-step exploration+all directions}}&&&&&&&&&&\\
\hline
Full model&{\color{ggray}\textit{as most}} 1-step exploration & 0.66 & 3.72 & 12.2 & 0.76 & 0.53 & 0.55 & 4.82 & 13.7 & 0.66 & 0.42 \\
(\S\ref{sec:mde})&{\color{ggray}\textit{at most}} 3-step exploration & 0.68 & 3.21 & 17.3 & 0.79 & 0.52 & 0.57 & 4.50 & 18.6 & 0.69 & 0.40 \\
{\color{ggray}\textit{parts of}}&{\color{ggray}\textit{at most}} \textbf{4-step exploration} & 0.70 & 3.20 & 19.7 & 0.80 & 0.52 & 0.58 & 4.36 & 20.6 & 0.70 & 0.40 \\
{\color{ggray}\textit{directions}}&{\color{ggray}\textit{at most}} 6-step exploration & 0.70 & 3.13 & 22.7 & 0.83 & 0.49 & 0.58 & 4.21 & 23.6 & 0.73 & 0.38  \\
\hline
\end{tabular}
}
\end{center}
\end{table}

\subsection{Diagnostic Experiments}
\label{sec:abs}

\noindent\textbf{Effectiveness of Our Basic Idea.} We first examine the performance of the na\"{i}ve model (\S\ref{sec:native}). As shown in Table\!~\ref{table:abs}, even with a simple exploration ability, the agent gains significant improvements over SR, NE and OR. It is no surprise to see drops in TL and SPL, as the agent simply explores all directions.

\noindent\textbf{Exploration Decision Making.} In \S\ref{sec:wte}, the agent learns to select some valuable directions to explore. As seen, the improved agent is indeed able to collect useful surrounding information by only conducting necessary exploration, as TL and SPL are improved without sacrificing improvements in SR, NE and OR.

\noindent\textbf{Allowing Multi-Step Exploration.} In \S\ref{sec:mde}, instead of only allowing one-step exploration, the agent learns to conduct multi-step exploration. To investigate the efficacy of such a strategy individually, we allow our na\"{i}ve model to make at most 4-step exploration (\textit{w/o.} exploration decision making). In Table\!~\ref{table:abs}, we can observe further improvements over SR, NE and OR scores, with larger TL.

\noindent\textbf{Importance of All Components.}  Next we study the efficacy of our full model from \S\ref{sec:mde}, which is able to make multi-direction, multi-step exploration. We find that, by integrating all the components together, our agent with at most 4-step exploration achieves the best performance in most metrics.

\noindent\textbf{Influence of Maximum Allowable Exploration Step.} From Table~\ref{table:abs}, we find that, with more maximum allowable exploration steps (1$\rightarrow$4), the agent attains better performance. However, allowing further exploration steps (4$\rightarrow$6) will hurt the performance. For at most 4-step exploration, the average exploration rate is 15.3\%. During exploration, the percentage of wrong navigation actions being corrected is $\sim$65.2\%, while right navigation action being changed wrongly is $\sim$10.7\%. The percentages of maximum exploration steps, from 1 to 4, are 53.6\%, 12.5\%, 8.7\%, and 25.3\%, respectively. We find that, in most cases, one-step exploration is enough. Sometimes the agent may choose long exploration, which maybe because he needs to collect more information for hard examples.



\begin{figure}[t]
      \centering
          \includegraphics[width=0.99\linewidth]{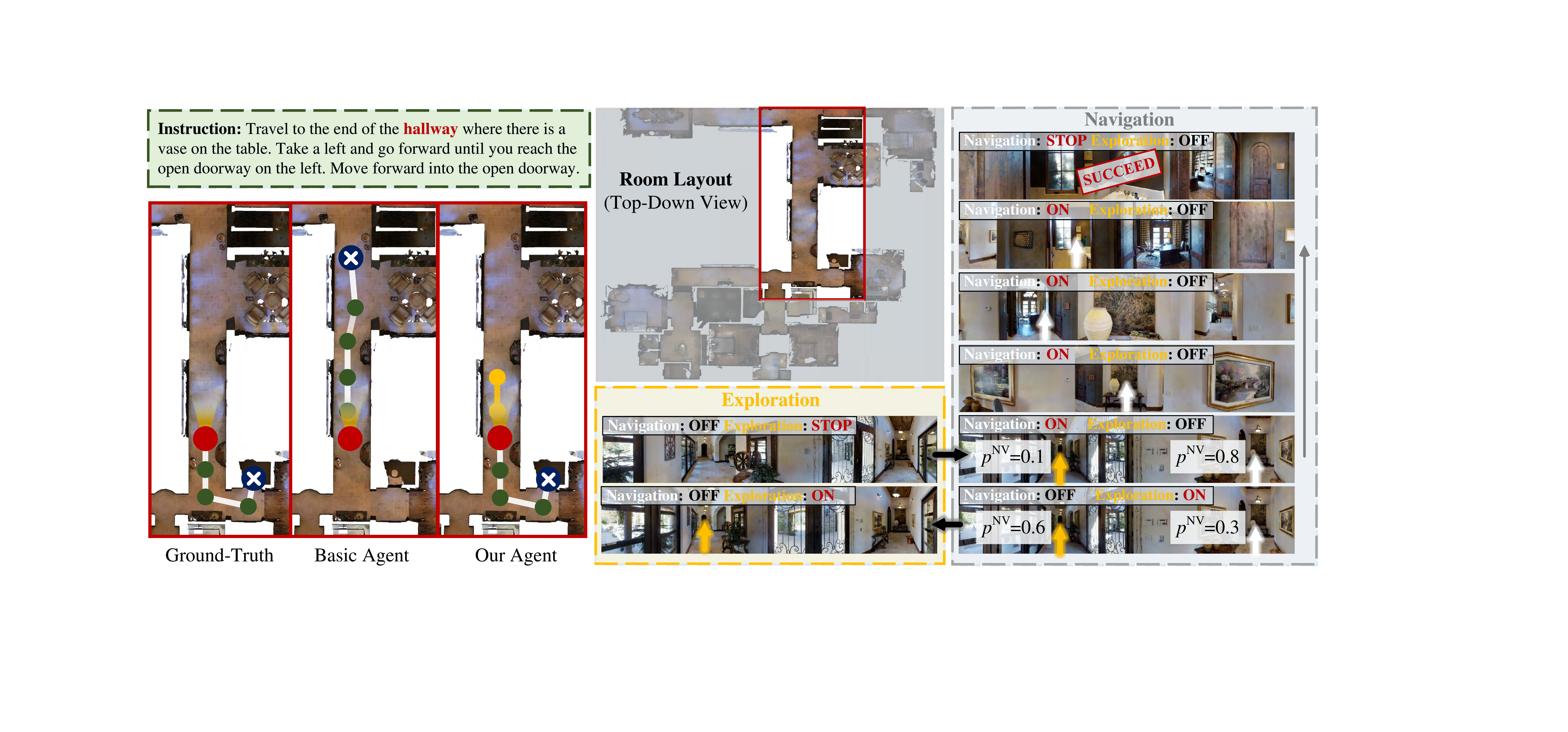}
    \caption{{\textbf{Left:} The basic agent is confused by the ambiguous instruction ``\textit{Travel to the end of the hallway}$\cdots$'', causing failed navigation. Our agent can actively collect information (the yellow part) and then make a better navigation decision. \textbf{Middle Bottom}: First view during exploration. \textbf{Right:} First view during navigation. We can find that, before exploration, the wrong direction gains a high navigation probability (\ie, 0.6). However, after exploration, the score for the correct direction is improved.
    }
    }
    \label{fig:vresults}
\end{figure}

\noindent\textbf{Qualitative Results.} Fig.\!~\ref{fig:vresults} depicts a challenge example, with the ambiguous instruction ``\textit{Travel to the end of the hallway}$\cdots$''. The basic agent chooses the wrong direction and ultimately fails. However, our agent is able to actively explore the environment and collect useful information, to support the navigation-decision making.  We observe that, after exploration, the correct direction gains a significant score and our agent reaches the goal location successfully.

\section{Conclusion}
This work proposes an end-to-end trainable agent for the VLN task, with an active exploration ability. The agent is able to intelligently interact with the environment and actively gather information when faced with ambiguous instructions or unconfident navigation decisions.
The elaborately designed exploration module successfully learns its own policy with the purpose of supporting better navigation-decision making. Our agent shows promising results on R2R dataset.

{\noindent\textbf{Acknowledgements}  This work was partially supported
by Natural Science Foundation of China (NSFC) grant (No.61472038), Zhejiang Lab's Open Fund (No.~2020AA3AB14), Zhejiang Lab's International Talent$_{\!}$ Fund$_{\!}$ for$_{\!}$ Young$_{\!}$ Professionals, and Key Laboratory of Electronic Information Technology in Satellite Navigation (Beijing Institute of Technology), Ministry of Education, China.}

\bibliographystyle{splncs04}
\bibliography{egbib}



\newpage

\appendix
\setcounter{table}{0}
\setcounter{figure}{0}
\renewcommand{\thetable}{A\arabic{table}}
\renewcommand{\thefigure}{A\arabic{figure}}

\begin{center}
    \Large{\textbf{Supplementary Material}}
\end{center}

\blfootnote{\Letter~Corresponding author: \textit{Wenguan Wang} (wenguanwang.ai@gmail.com).}

In this document, we first detail our memory based late action-taking strategy in \S\ref{sec:MLA}. Then, we present the work and data flow of our navigation agent (see \S\ref{sec:WF}). Later, in \S\ref{sec:DE}, more ablation studies are conducted to fully assess the effectiveness of our learning protocol. Later, in~\S\ref{sec:qa}, visual results for some representative successful and failed cases as well as analyses are presented.  Finally, in \S\ref{sec:statistics}, we provide more statistics of our exploration module, to give an in-depth glimpse into our agent.

\section{Memory based Late Action-Taking Strategy}
\label{sec:MLA}
As described above, after a round of exploration, the agent goes back to the starting point at $t^{th}$ navigation step to update the knowledge collected about a certain direction. However, this "going back" may result in the repeated visiting of viewpoints in three manners, 1) the re-visiting of the viewpoints on the way back to the starting point, 2) the re-visiting of the viewpoints within the rounds of exploration at $t^{th}$ navigation step. Though each exploration leads to a certain direction (viewpoint), those directions may interact somewhere in front and the intersection may be re-visited. 3) the re-visiting of the viewpoints across different navigation steps, \eg, the viewpoints explored at $t^{th}$ navigation step may be re-visited at $t+1^{th}$ navigation step. Those re-visiting will lead to additional TL.

To shrink the trajectory, a novel moving strategy is adopted. We call it ``lazy'' strategy. The core idea of the strategy is that the agent moves only when it is necessary. Specifically, supposing that the agent finished a round exploration at viewpoint $v_t$. We store the views of all the visited points and the connectivity between the points in a memory graph, so the agent can ``image'' the execution of his following action without really going back to the starting point. When he needs to visit a point (assumed as $v$) that is not stored in the memory graph, the new point $v$ will be added to the memory graph and the agent will go to $v$ from $v_t$ following the shortest path in the memory graph, then the view of $v$ will be stored. The agent will not leave $v$ until he needs to visit a new point that is not stored in the memory graph or he decides to stop the navigation. In this way, a lot of repeated visits can be saved.

\section{Work and Data Flow}
\label{sec:WF}
\begin{figure}[t]
      \centering
          \includegraphics[width=0.9\linewidth]{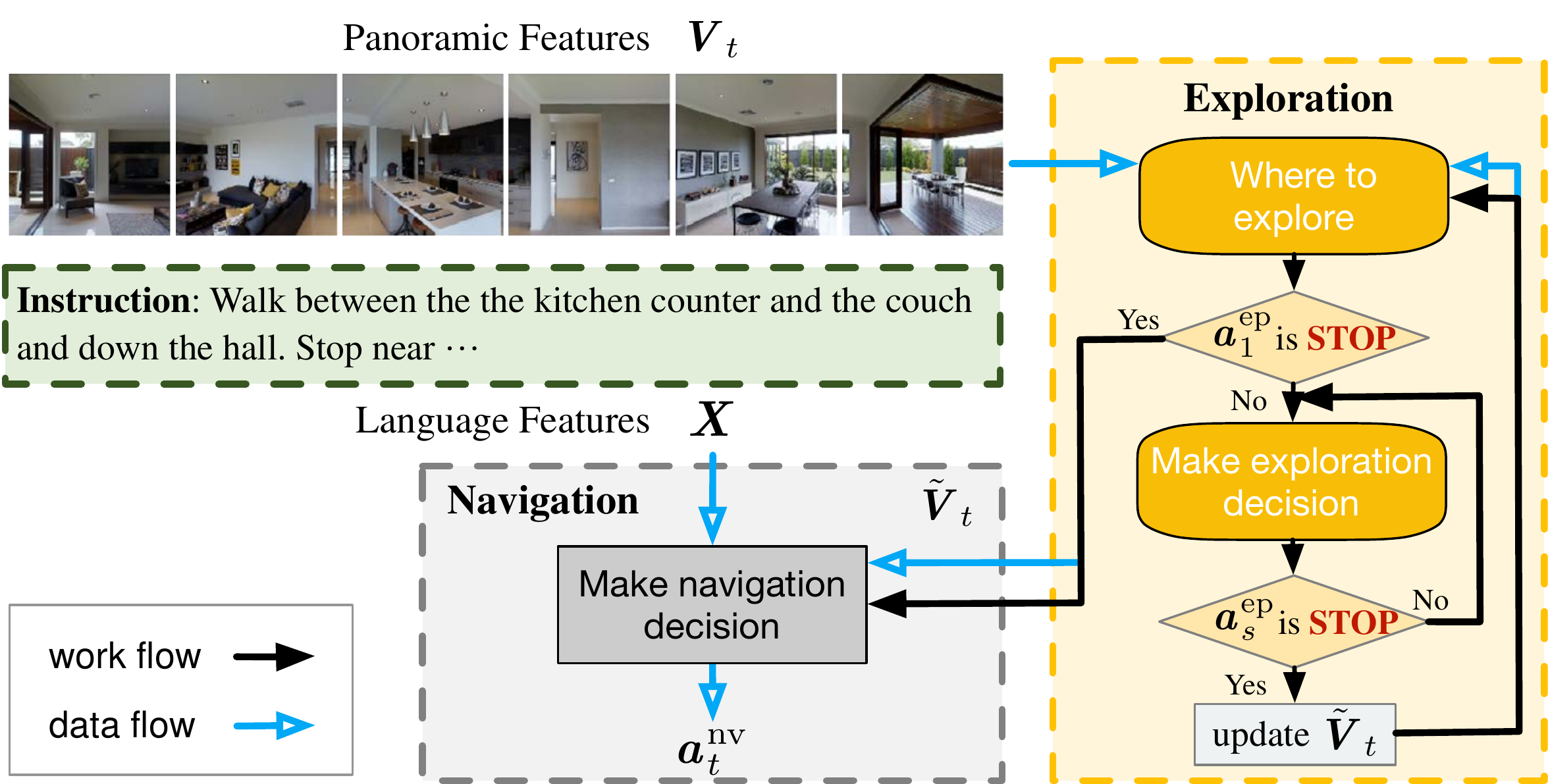}
    \caption{The chart of work flow and data flow at navigation step $t$. }
    \label{fig:dataflow}
\end{figure}

 Fig.~\ref{fig:dataflow} depicts our work and data flow during $t^{th}$ step navigation. At first, the exploration module decides where to explore. Supposing $k$ direction is selected, the agent continues to make exploration decisions and collects surrounding visual information, until \texttt{STOP} action is selected. Then he returns to $t^{th}$  navigation point, updates the visual feature of $k$ direction, and prepares for next-round exploration. If the exploration module make a stop decision, the updated visual feature is used to make a navigation decision. Otherwise he will select a direction to make exploration.

\section{Diagnostic Experiments of Training Signals}
\label{sec:DE}

In this section, we make diagnostic experiments to analyze the effectiveness of our training signals. Our agent is trained with two distinct learning paradigms, \ie, (1) imitation learning, and (2) reinforcement learning. Our basic agent~\cite{tan2019learning} has examined the effectiveness of the reinforcement learning loss $\mathcal{L}^{\text{nv}}_{\text{RL}}$ for the whole network training. Here we design the experiments to evaluate the effectiveness of the imitation learning loss $\mathcal{L}^{\text{ep}}_{\text{IL}}$ and the reinforcement learning loss $\mathcal{L}^{\text{ep}}_{\text{RL}}$ used during the training of our exploration module. Specifically, we retrain our agent (at most 4-step exploration) without either $\mathcal{L}^{\text{ep}}_{\text{IL}}$ or $\mathcal{L}^{\text{ep}}_{\text{RL}}$ in the final loss.

As shown in Table~\ref{table:abs2}, with either imitation learning or reinforcement learning signal, our full-model achieves higher \textbf{SR} compared to the basic agent, which is improved by 1\% and 3\% on validation unseen set respectively. This shows the effectiveness of two training signals. Note that the agent with the mixture of $\mathcal{L}^{\text{ep}}_{\text{IL}}+\mathcal{L}^{\text{ep}}_{\text{RL}}$ outperforms the one without $\mathcal{L}^{\text{ep}}_{\text{RL}}$ or $\mathcal{L}^{\text{ep}}_{\text{IL}}$ by 5\% and 3\% on validation unseen respectively. It means that the mixture signal somewhat overcomes the relative poor generalizability of IL and difficult convergence of RL.

\begin{table}[t]
    \begin{center}
        \caption{{Ablation study about training signals on \texttt{validation} \texttt{seen} and \texttt{validation} \texttt{unseen} sets of R2R dataset~\cite{anderson2018vision} under \textbf{Single Run} setting.}}
    \label{table:abs2}
            \resizebox{1\textwidth}{!}{
            \setlength\tabcolsep{1pt}
            \renewcommand\arraystretch{1.1}
    \begin{tabular}{c|c||ccccc|ccccc}
    \hline \thickhline
    \rowcolor{mygray}
    & & \multicolumn{10}{c}{Single Run Setting} \\
    \cline{3-12}
    \rowcolor{mygray}
    & & \multicolumn{5}{c|}{\texttt{validation} \texttt{seen}} & \multicolumn{5}{c}{\texttt{validation} \texttt{unseen}} \\
    \cline{3-12}
    \rowcolor{mygray}
    \multicolumn{1}{c|}{\multirow{-3}{*}{Model}} &\multirow{-3}{*}{Loss Component} &\textbf{SR}$\uparrow$ &NE$\downarrow$ &TL$\downarrow$ &OR$\uparrow$ &SPL$\uparrow$ &\textbf{SR}$\uparrow$ &NE$\downarrow$ &TL$\downarrow$ &OR$\uparrow$  &SPL$\uparrow$\\
    \hline
    basic agent &$\mathcal{L}^{\text{nv}}_{\text{IL}}+\mathcal{L}^{\text{nv}}_{\text{RL}}$ &0.62 & 3.99 & 11.0 & 0.71 & 0.59 & 0.52 & 5.22 & 10.7 & 0.58 &0.48 \\

    \hline
    &w/o. $\mathcal{L}^{\text{ep}}_{\text{IL}}$ &\multirow{2}{*}{0.65} &\multirow{2}{*}{3.70} &\multirow{2}{*}{18.3} &\multirow{2}{*}{0.76} &\multirow{2}{*}{0.53} &\multirow{2}{*}{0.53} &\multirow{2}{*}{5.15} &\multirow{2}{*}{17.1} &\multirow{2}{*}{0.65} &\multirow{2}{*}{0.39} \\
    \specialrule{0em}{-0.5pt}{-1.5pt}
    &{\color{ggray}\textit{$\mathcal{L}^{\text{nv}}_{IL}+\mathcal{L}^{\text{nv}}_{RL}+\mathcal{L}^{\text{ep}}_{\text{RL}}$}} &&&&&&&&&&\\
    &w/o. $\mathcal{L}^{\text{ep}}_{\text{RL}}$&\multirow{2}{*}{0.67} &\multirow{2}{*}{3.53} &\multirow{2}{*}{17.6} &\multirow{2}{*}{0.79} &\multirow{2}{*}{0.51} &\multirow{2}{*}{0.55} &\multirow{2}{*}{4.65} &\multirow{2}{*}{18.9} &\multirow{2}{*}{0.68} &\multirow{2}{*}{0.39} \\
    \specialrule{0em}{-0.5pt}{-1.5pt}
    &{\color{ggray}\textit{$\mathcal{L}^{\text{nv}}_{IL}+\mathcal{L}^{\text{nv}}_{RL}+\mathcal{L}^{\text{ep}}_{\text{IL}}$}}&&&&&&&&&&\\
    &full loss &\multirow{2}{*}{0.70} &\multirow{2}{*}{3.20} &\multirow{2}{*}{19.7} &\multirow{2}{*}{0.80} &\multirow{2}{*}{0.52} &\multirow{2}{*}{0.58} &\multirow{2}{*}{4.36} &\multirow{2}{*}{20.6} &\multirow{2}{*}{0.70} & \multirow{2}{*}{0.40}\\
    \specialrule{0em}{-0.5pt}{-1.5pt}
    \multirow{-6}{*}{full model}&{\color{ggray}\textit{$\mathcal{L}^{\text{nv}}_{IL}+\mathcal{L}^{\text{nv}}_{RL}+\mathcal{L}^{\text{ep}}_{\text{IL}}+\mathcal{L}^{\text{ep}}_{\text{RL}}$}}&&&&&&&&&&\\
    \hline
    w/o. ``Lazy'' strategy & full loss &0.70 & 3.20 & 45.3 & 0.80 & 0.29 & 0.58 & 4.36 & 44.5 & 0.70 &0.18 \\
    \hline
    \end{tabular}
    }
    \end{center}
\end{table}

\section{Qualitative Analysis}
\label{sec:qa}
In this section, we visualize some successful and failed cases of our agent and further provide in-depth analyses.
\subsection{Successful Examples}
We show three representative examples, over which the basic agent fails while our agent succeeds.

\begin{figure}[!t]
      \centering
          \includegraphics[width=0.99\linewidth]{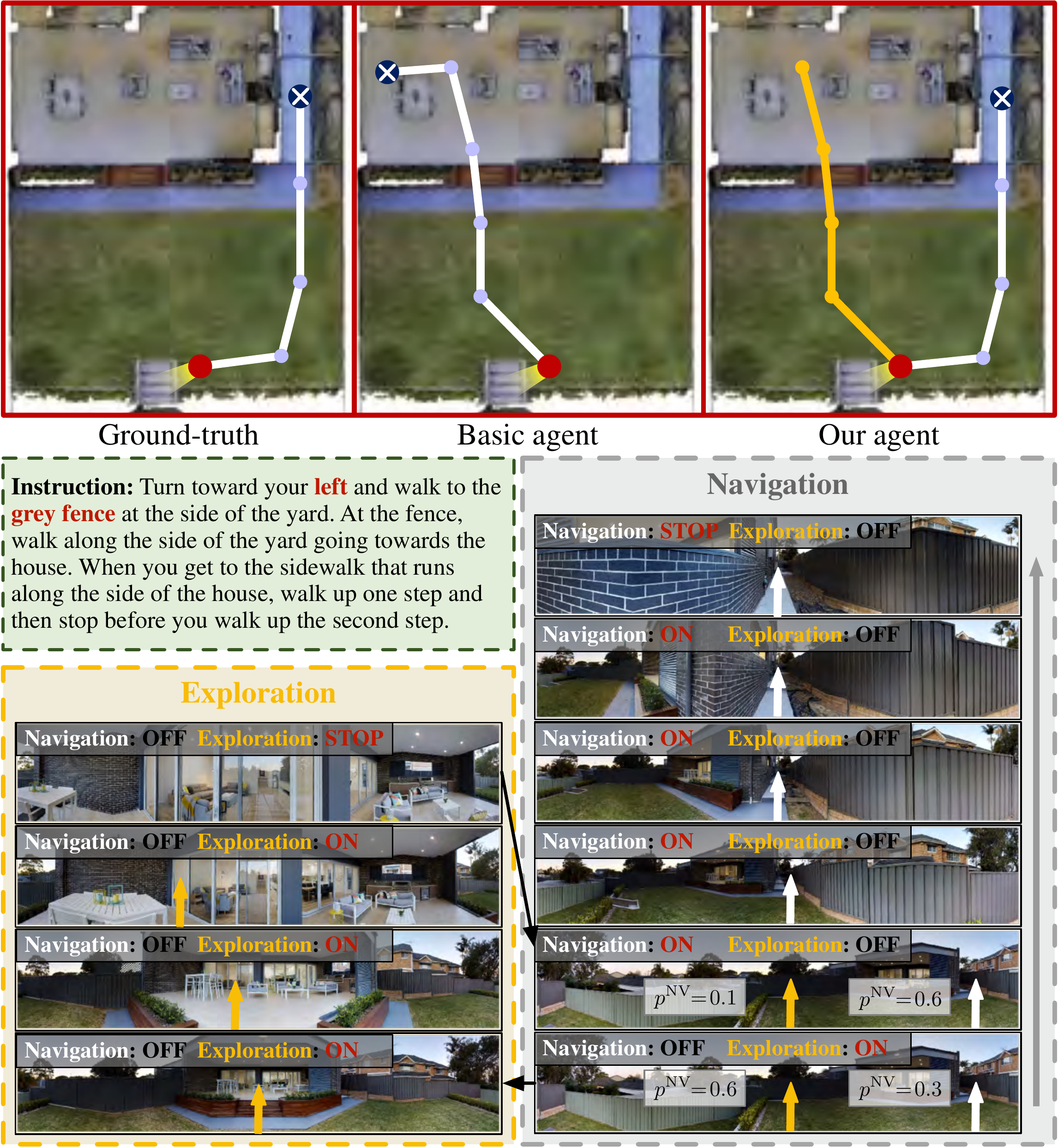}
    \caption{{Case 1. \textbf{Top:} Groundtruth navigation route,  basic agent's navigation route, and our agent's exploration and navigation routes. Note that the initial orientation of the agent is highlighted. The basic agent gets confused with the wrong instruction ``\textit{Turn toward your left}$\cdots$''. In addition,  there exist different ways to the ``\textit{grey fence}''. Thus he fails finally. However, our agent can actively explore the surrounding (the yellow part) and then make more accurate navigation decision. \textbf{Bottom Left}: First view during exploration of our agent. \textbf{Bottom Right:} First view during navigation of our agent. We can find that, before making exploration, the wrong direction gains a high navigation probability (\ie, 0.6). However, after exploration, the probability for correct direction is improved (\ie, $0.3$$\rightarrow$$0.6$). This case is taken from R2R dataset\cite{anderson2018vision}, path 3492, instruction set 2.
    }
    }
    \label{fig:case1}
\end{figure}

\begin{figure}[!t]
      \centering
          \includegraphics[width=0.99\linewidth]{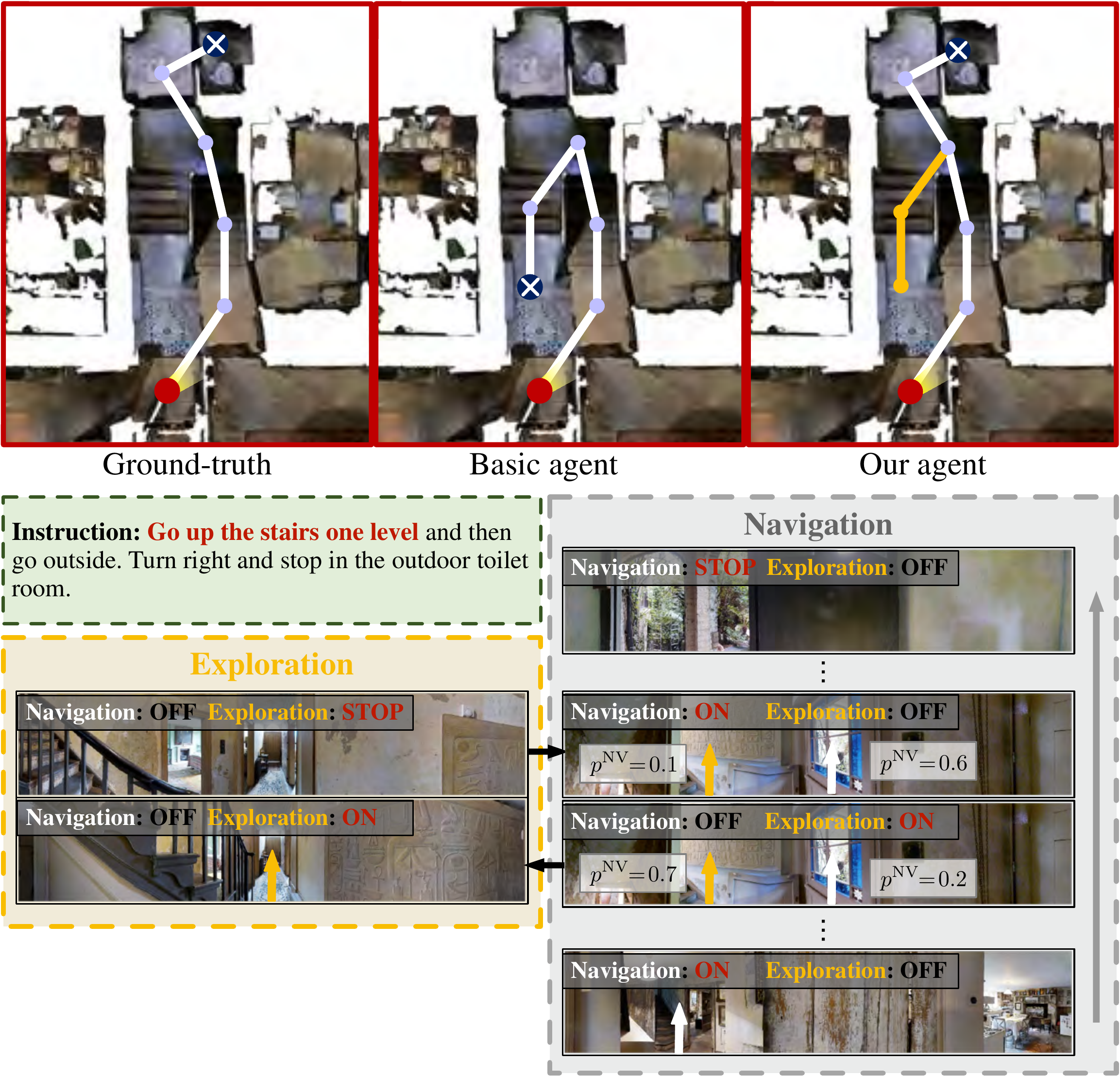}
    \caption{{Case 2. \textbf{Top:} Groundtruth navigation route,  basic agent's navigation route, and our agent's exploration and navigation routes. Note that the initial orientation of the agent is highlighted.  The basic agent is hard to understand the instruction ``\textit{Go up the stairs one level}$\cdots$'' and thus continues to climb another staircase. For our agent, after collecting more information, a correct navigation direction is chose and reaches the goal successfully. \textbf{Bottom Left}: First view during exploration of our agent. Our agent explores the upstairs and finds the direction is wrong. \textbf{Bottom Right:} First view during navigation of our agent. After the exploration, the probability of the correct direction is improved (\ie, $0.2$$\rightarrow$$0.6$). This case is taken from R2R dataset\cite{anderson2018vision}, path 3766, instruction set 3.}
    }
    \label{fig:case2}
\end{figure}

\begin{figure}[!t]
      \centering
          \includegraphics[width=0.99\linewidth]{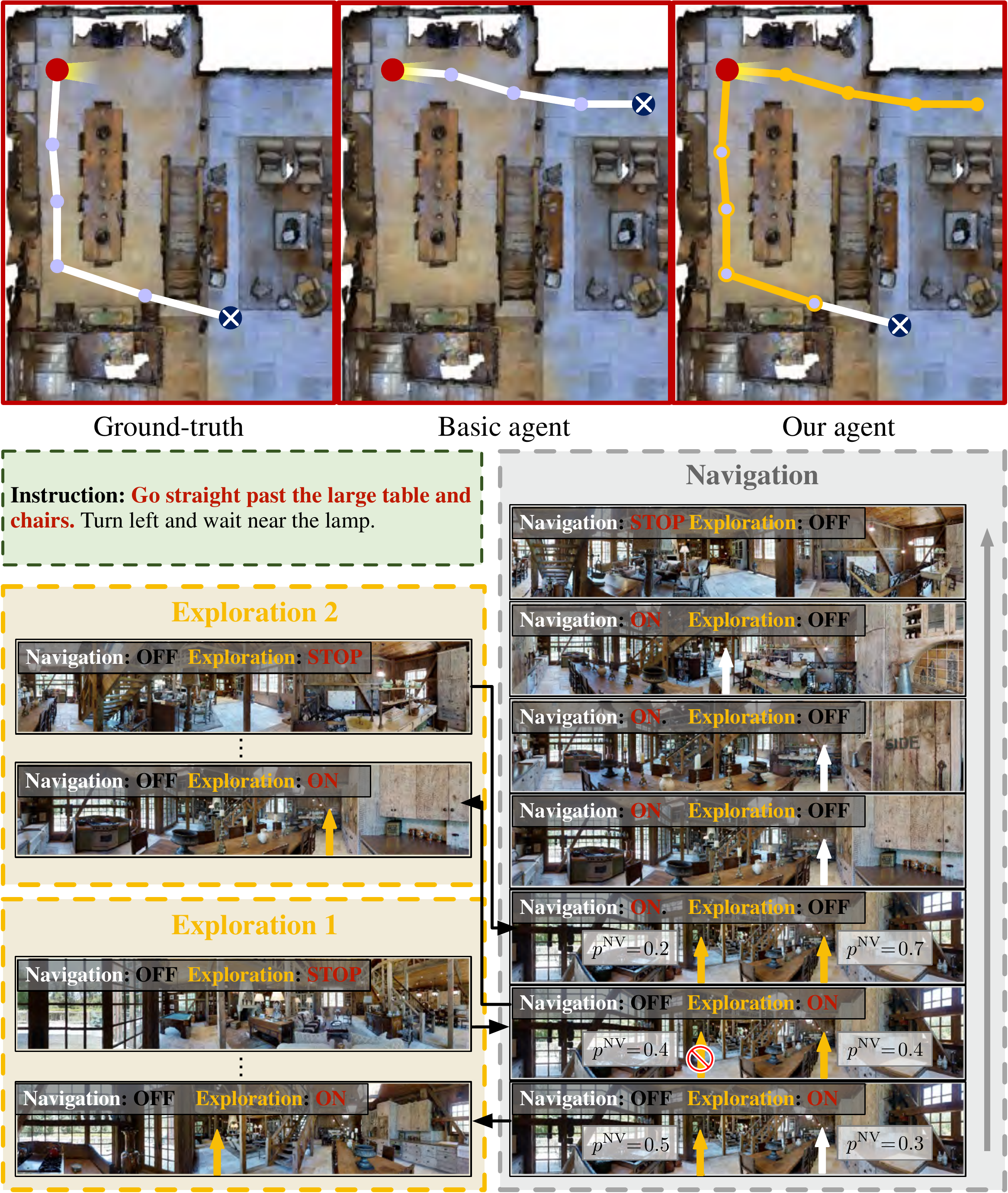}
    \caption{{Case 3. \textbf{Top:} Groundtruth navigation route,  basic agent's navigation route, and our agent's exploration and navigation routes. Note that the initial orientation of the agent is highlighted.  The instruction ``\textit{Go straight past the large table and chairs.}$\cdots$'' is wrong, as the instructor does not take notice of the agent's initial orientation.  \textbf{Bottom Left}: First view of our agent during two exploration rounds. \textbf{Bottom Right:} First view during navigation of our agent. We can find that, after first-round exploration, the probability of the wrong direction is suppressed (\ie, $0.5$$\rightarrow$$0.4$). After the second round of exploration, the probability for correct direction is improved (\ie, $0.4$$\rightarrow$$0.7$). This case is taken from R2R dataset\cite{anderson2018vision}, path 4782, instruction set 1.
    }
    }
    \label{fig:case3}
\end{figure}

\begin{figure}[!t]
      \centering
          \includegraphics[width=0.99\linewidth]{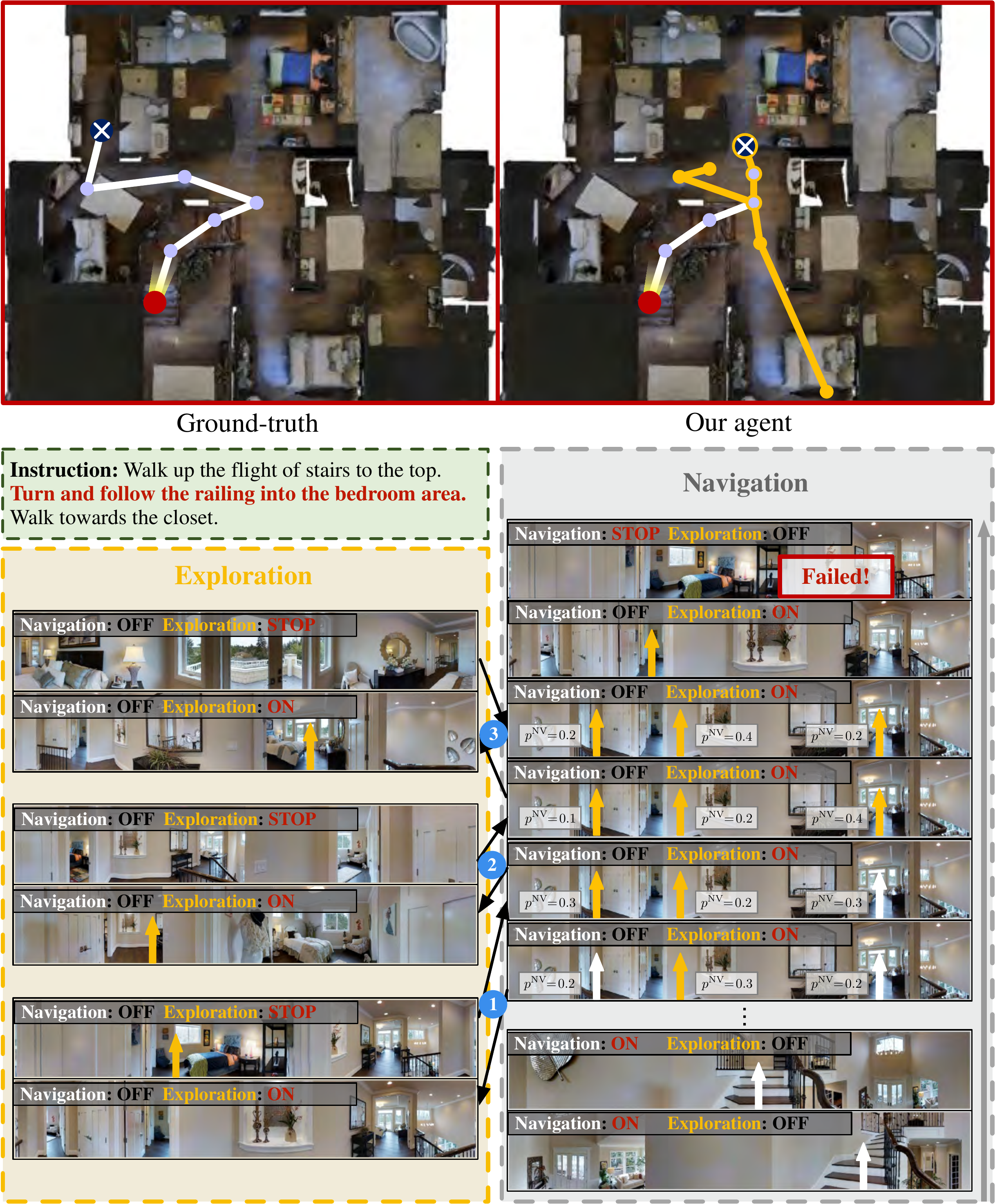}
    \caption{{Failed Case. \textbf{Top:} Groundtruth navigation route,  basic agent's navigation route, and our agent's exploration and navigation routes.  Note that the initial orientation of the agent is highlighted.   \textbf{Bottom Left}: First view during three rounds of exploration by our agent. \textbf{Bottom Right:} First view during navigation of our agent. Our agent conducts three rounds  of exploration at the top of the stairs. He fails to collect enough information during the second-round exploration, causing failed navigation.
    This case is taken from R2R dataset\cite{anderson2018vision}, path 21, instruction set 1.
    }
    }
    \label{fig:fail}
\end{figure}

For the first case in Fig.~\ref{fig:case1}, the instruction, ``\textit{Turn toward your left}'', is inconsistent with the initial orientation of the agent. Additionally, the important landmark, ``\textit{grey fence}'', can be observed from two different directions. These issues result in the confusion of the basic agent and unluckily lead to failure. Our agent conducts the active exploration to mitigate the ambiguity of the instruction and finally chooses the right route.

The second case is shown in Fig.~\ref{fig:case2}. In this case, the instruction said ``\textit{Go up the stairs one level}$\cdots$'', where ``\textit{one level}'' has ambiguous definition in the dataset as the routes are annotated by different workers. The basic agent continues to climb another staircase and fails finally. After the exploration, our agent finds that after climbing another staircase, he cannot execute the following instructions. Then the agent chooses the right direction.

In Fig.~\ref{fig:case3}, the initial instruction is ambiguous, as the visual context of both directions are consistent with the instruction ``$\cdots$ \textit{past the large table and chairs.}''. Moreover, if considering the mismatch between the instruction ``\textit{Go straight} $\cdots$'' and the agent's initial orientation, the wrong direction even looks more consistent with the instruction. After two rounds of exploration, our agent finds that the last half routes of the exploration in the right direction is more consistent with the following instructions, which is more related to the goal viewpoint comparing to the starting instruction. Then our agent chooses the right direction.

\subsection{Failed Examples}
In this section, we show a representative failed example, which also demonstrates possible directions of future efforts.

As shown in Fig.~\ref{fig:fail}, our agent first correctly executes the instruction ``\textit{Walk up the flight of stairs to the top.}''. At the top of the stairs, our agent is confused by the following instruction ``\textit{Turn and follow the railing into the bedroom area.}''. This because there are three directions to different bedrooms and two of them have railing along the way. Thus our agent decides to explore the surroundings.

During the first-round exploration, the agent explores the front door and finds this direction is wrong. Then the probability of this direction is suppressed (0.3$\rightarrow$0.2). Then, the agent selects a new direction and makes a second-round exploration. Though this direction is correct, the agent does not step into the correct room and misses the critical landmark, ``\textit{closet}''. Thus the probability of the correct direction is mistakenly suppressed (0.3$\rightarrow$0.1). In the third-round exploration, the agent explores the door on the right and finds that it is still inconsistent with the description of the final goal viewpoint. The probability of the last possible direction is suppressed again  (0.4$\rightarrow$0.2). Then our agent picks the most plausible direction, which is of the highest navigation probability, but fails. We think there are two possible reasons for the failure of the exploration. First, our exploration module mainly focuses on current navigation action related instruction, but lacks the ability of dynamically parsing the following instructions during exploration. Maybe a multi-head co-attention or future navigation action aware exploration module is needed. Second, current navigation module may needs to be equipped with a reference grounding ability, thus the essential reference ``\textit{closet}'' will be addressed during exploration.

\section{Statistics Analysis}
\label{sec:statistics}
In this section, we provide more statistic analyses about our exploration module. The statistics are collected from our full agent with 4-step exploration module on the validation unseen split of R2R dataset \cite{anderson2018vision} .

The overall average exploration rate is 15.3\%, \ie, the probabilities of making exploration during a navigation step is 15.3\%.  When the agent decides to explore, the average number of exploration directions is 1.82, the average, pre-direction exploration steps are 2.05. The average trajectory length (\textbf{TL}) is 20.6, in which the length of navigation is 9.4, while the length of exploration is 11.2. We can find that the real navigation routes are very short. For the successful and failed navigation cases,
the average trajectory lengths are  18.7 and 23.2 respectively. \textbf{TL} of failed cases is longer by 4.5, which possibly means that our agent is confused in the failed cases and tries to collect more information through making more explorations.

We also make specific analysis on the effect of exploration by studying how the exploration influences the navigation action decisions. For the navigation steps with exploration, the rate that the original navigation action is changed through exploration is 75.1\%, in which the rate that wrong navigation action is corrected is about
86.5\%. That means in most cases our exploration  module could collect meaningful information for supporting navigation-decision making.


\input{./figs/appendix.brf}

\end{document}